\documentclass[lettersize,journal]{IEEEtran}
\pdfoutput=1

\usepackage{amsmath,amsfonts}
\usepackage{algorithmic}
\usepackage{algorithm}
\usepackage{array}
\usepackage[caption=false,font=normalsize,labelfont=sf,textfont=sf]{subfig}
\usepackage{textcomp}
\usepackage{stfloats}
\usepackage{url}
\usepackage{verbatim}
\usepackage{graphicx}
\usepackage{epstopdf}
\usepackage{cite}
\usepackage{makecell}
\usepackage{multirow}
\usepackage{colortbl}
\usepackage{booktabs,graphicx}
\usepackage{mathtools}
\usepackage{caption}
\usepackage{float}
\usepackage{}
\usepackage{bbding}

\usepackage{wrapfig}
\def\ie{\textit{i.e.}}

\usepackage[utf8]{inputenc} 
\usepackage[T1]{fontenc}    
\usepackage{hyperref}       
\usepackage{url}            
\usepackage{booktabs}       
\usepackage{amsfonts}       
\usepackage{nicefrac}       
\usepackage{microtype}      
\usepackage{xcolor}         
\usepackage{ulem}

\usepackage{booktabs}
\definecolor{red}{RGB}{255,0,0}
\definecolor{orange}{RGB}{255, 127, 0}
\newcommand{\blue}[1]{\textcolor{blue}{#1}}

\begin{document}
{\onecolumn

\noindent \vspace{1cm}

\noindent \textbf{\huge{DilateFormer:  Multi-Scale Dilated Transformer \\
\\
for Visual Recognition}}

\vspace{2cm}

\noindent {\LARGE{Jiayu Jiao, Yu-Ming Tang, Kun-Yu Lin, Yipeng Gao, Jinhua Ma, \\
\\
Yaowei Wang, Wei-Shi Zheng*}}
\\
\\
*Corresponding author: Wei-Shi Zheng.
\\
\\
Code: \href{https://github.com/JIAOJIAYUASD/dilateformer}{\blue{https://github.com/JIAOJIAYUASD/dilateformer}}
\\
\\
Project page: \href{https://isee-ai.cn/~jiaojiayu/DilteFormer.html}{\blue{https://isee-ai.cn/~jiaojiayu/DilteFormer.html}}

\vspace{1cm}

\noindent {\LARGE{Submission date: 22-Sep-2022 to IEEE Transaction on Multimedia
}}

\vspace{1cm}

\noindent For reference of this work, please cite:

\vspace{1cm}
\noindent Jiayu Jiao, Yu-Ming Tang, Kun-Yu Lin, Yipeng Gao, Jinhua Ma, Yaowei Wang, and Wei-Shi Zheng.
``DilateFormer: Multi-Scale Dilated Transformer
for Visual Recognition''. \emph{IEEE Transaction on Multimedia,} 2023.

\vspace{1cm}

\noindent Bib:\\
\noindent @article\{jiao2023dilateformer,\\
\ \ \  title     = \{DilateFormer: Multi-Scale Dilated Transformer for Visual Recognition\}, \\
 \ \ \   author    = \{Jiao, Jiayu and Tang, Yu-Ming and Lin, Kun-Yu and Gao, Yipeng and Ma, Jinhua and Wang, Yaowei and Zheng, Wei-Shi\},\\
\ \ \  journal   = \{\{IEEE\} Transaction on Multimedia\},\\
\ \ \  year      = \{2023\}\\
\}
}

{
\twocolumn

\title{DilateFormer:  Multi-Scale Dilated \\ Transformer 
for Visual Recognition}

\author{Jiayu Jiao,
Yu-Ming Tang, 
Kun-Yu Lin, 
Yipeng Gao, 
Jinhua Ma, 
Yaowei Wang,
Wei-Shi Zheng*
\thanks{\scriptsize J. Jiao and Y.-M. Tang are equally-contributed authors. *Corresponding author: W.-S. Zheng.}
\thanks{\scriptsize J. Jiao, Y.-M. Tang, K.-Y. Lin, Y. Gao and J. Ma are with the School of Computer Science and Engineering, Sun Yat-Sen University, Guangzhou, China (e-mail: \{jiaojy6, tangym9, linky5, gaoyp23\}@mail2.sysu.edu.cn, majh8@mail.sysu.edu.cn).}
\thanks{\scriptsize W.-S. Zheng is with the School of Computer Science and Engineering, Sun Yat-sen University, Guangzhou, China, with Peng Cheng Laboratory, Shenzhen, China, and also with the Key Laboratory of Machine Intelligence and Advanced Computing (Sun Yat-sen University), Ministry of Education, China. (e-mail: wszheng@ieee.org /zhwshi@mail.sysu.edu.cn).}
\thanks{\scriptsize Y. Wang is with the Pengcheng Laboratory, ShenZhen, China (e-mail: wangyw@pcl.ac.cn).}
}

\markboth{SUBMISSION TO IEEE TRANSACTIONS ON MULTIMEDIA}%
{Shell \MakeLowercase{\textit{et al.}}: Bare Demo of IEEEtran.cls for IEEE Journals}


\maketitle

\begin{abstract}
As a \textit{de facto} solution, the vanilla Vision Transformers (ViTs) are encouraged to model long-range dependencies between arbitrary image patches while the global attended receptive field leads to quadratic computational cost. Another branch of Vision Transformers exploits local attention inspired by CNNs, which only models the interactions between patches in small neighborhoods. Although such a solution reduces the computational cost, it naturally suffers from small attended receptive fields, which may limit the performance. 
In this work, we explore effective Vision Transformers to pursue a preferable trade-off between the computational complexity and size of the attended receptive field. By analyzing the patch interaction of global attention in ViTs, we observe two key properties in the shallow layers, namely locality and sparsity, indicating the redundancy of global dependency modeling in shallow layers of ViTs. Accordingly, we propose Multi-Scale Dilated Attention (MSDA) to model \textit{local} and \textit{sparse} patch interaction within the sliding window. With a pyramid architecture, we construct a Multi-Scale Dilated Transformer (DilateFormer) by stacking MSDA blocks at low-level stages and global multi-head self-attention blocks at high-level stages. Our experiment results show that our DilateFormer achieves state-of-the-art performance on various vision tasks. On ImageNet-1K classification task, DilateFormer achieves comparable performance with 70\% fewer FLOPs compared with existing state-of-the-art models.
Our DilateFormer-Base achieves 85.6\% top-1 accuracy on ImageNet-1K classification task, 53.5\% box mAP/46.1\% mask mAP on COCO object detection/instance segmentation task and 51.1\% MS mIoU on ADE20K semantic segmentation task.
\end{abstract}

\begin{IEEEkeywords}
Vision Transformer.
\end{IEEEkeywords}

\section{Introduction}
\label{intro}
\IEEEPARstart{I}{n} the past years, Convolution Neural Networks (CNNs) have dominated a wide variety of vision tasks such as classification \cite{AlexNet,vgg,resnet,googlenet,liu2022ConvNeXt, yixing_tmm,ZhaoXBGXD21}, object detection \cite{fastrcnn,yolov3,ssd,retinanet,he2017mask} and semantic segmentation \cite{unet,DeepLabV3,fcn},
attributing to the inductive bias of convolution operations, \ie, local connections and weight sharing. However, convolution only models local dependencies of pixels, which ignores the dependency modeling between distant pixels to some extent \cite{wang2018nonlocal}. Inspired by sequence modeling tasks \cite{Brown2020Language,radford2018improving} in natural language processing (NLP)\cite{Vaswani2017atten,radford2018improving,Beltagy2020Longformer}, pioneer works \cite{dosovitskiy2020image,deit,Touvron2021deeper,chu2021cpe,MaGSMKCTYF21} introduce Transformers with long-range dependency modeling ability into computer vision, achieving exciting results in various vision tasks.

\begin{figure}[t]
\centering
    \vspace{-2.5em}
    \includegraphics[width= 1.1\linewidth]{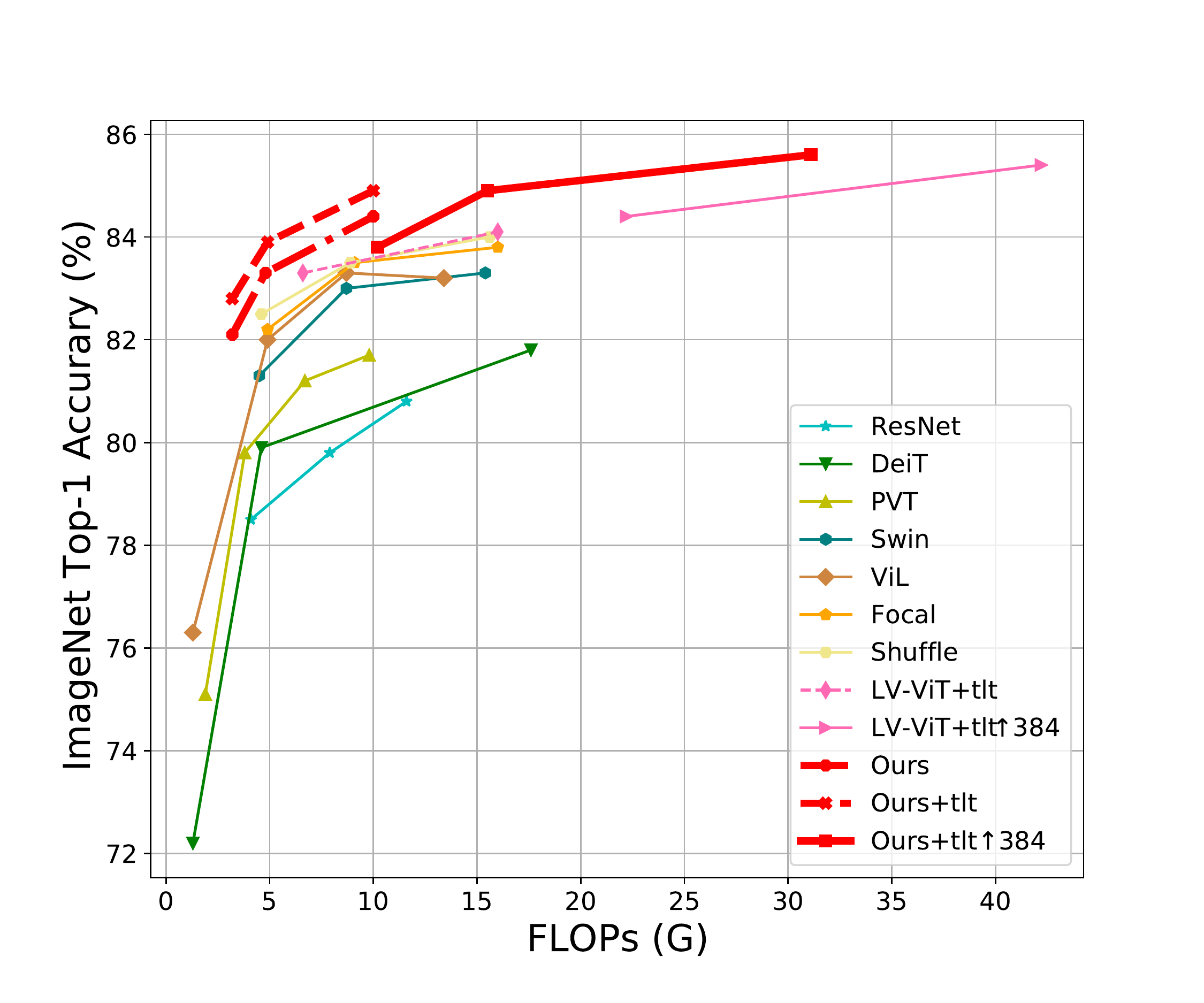}
    \vspace{-3em}
    \caption{Performance comparisons with respect to FLOPs on ImageNet-1K classification. Without extra training data, our DilateFormer variants achieve comparable or even better performance with fewer FLOPs.}
    \vspace{-2em}
    \label{fig:flop_acc1}
\end{figure}

With global attention, the vanilla Vision Transformers (ViTs)  \cite{dosovitskiy2020image,deit} can conduct dependency modeling between arbitrary image patches. However, the \textbf{global} attended receptive field of ViTs leads to quadratic computational cost, and modeling dependencies among all patches may be redundant for mainstream vision tasks. To reduce the computational cost and redundancy of global attention, some works \cite{liu2021swin,yang2021focal,zhang2021vil,chu2021twins,Hassani2022nat} introduce inductive bias explored in CNNs, performing \textbf{local} attention only in small neighborhoods. However, local attention naturally suffers from small attended receptive fields, which results in a lack of capability to model long-range dependencies. 

In this work, we explore an effective Vision Transformer to pursue a preferable trade-off between the computational complexity and the size of the attended receptive field. By analyzing the patch interaction of global attention in ViTs \cite{dosovitskiy2020image,deit}, we find that the attention matrix in shallow layers has two key properties, namely \textit{locality} and \textit{sparsity}. 
As shown in Figure \ref{atten_fig}, in the third attention block of ViT-Small, relevant patches are sparsely distributed in the neighborhood of the query patch. Such a locality and sparsity property indicates that distant patches in shallow layers are mostly irrelevant in semantics modeling for mainstream vision tasks, and thus there is much redundancy to be reduced in the costly global attention module.

Based on the above analysis, we propose a Sliding Window Dilated Attention (SWDA) operation, which performs self-attention among patches sparsely selected in the surrounding field. To make further use of the information within the attended receptive field, we propose  Multi-Scale Dilated Attention (MSDA), which simultaneously captures semantic dependencies at different scales.
MSDA sets different dilation rates for different heads, enabling the ability of multi-scale representation learning. 
Following PVT \cite{wang2021pyramid} and Swin \cite{liu2021swin},
we adopt a pyramid architecture to develop a new effective Transformer model, namely Multi-Scale Dilated Transformer (DilateFormer), which stacks MSDA in shallow stages to capture low-level information and global Multi-Head Self-Attention  \cite{dosovitskiy2020image,deit} in deeper stages to model high-level interaction.

For model evaluation, we design variants of DilateFormer with different capacities and apply them to different vision tasks. Experimental results show that our proposed DilateFormer outperforms state-of-the-art Vision Transformers \cite{deit,dosovitskiy2020image,liu2021swin,zhang2021vil,Hassani2022nat,yang2021focal} on various datasets across different model sizes. As depicted in Figure \ref{fig:flop_acc1}, we demonstrate the performance
of our DilateFormers on ImageNet-1K classification task. Without extra training data, our Dilate-S (4.8 GFLOPs) achieves comparable performance with Swin-B (15.4 GFLOPs) \cite{liu2021swin} on ImageNet-1K using only 1/3 FLOPs. With the assistance of Token Labeling
\cite{jiang2021tlt}, our DilateFormers achieve better performance than LV-ViTs \cite{jiang2021tlt} at different model sizes. Specifically, our Dilate-S$^{\star}$ (4.9 GFLOPs) and our Dilate-B$^{\star}$ (10.0 GFLOPs) achieve 83.9\% and 84.9\% respectively, surpassing LV-ViT-S  \cite{jiang2021tlt} (6.6 GFLOPs) and LV-ViT-M \cite{jiang2021tlt}(16 GFLOPs). 
Besides, our Dilate-B achieves 85.6\% top-1 accuracy on ImageNet-1K classification \cite{deng2009large} task, 53.5\% box mAP/46.1\% mask mAP on COCO \cite{lin2014microsoft} object detection/instance segmentation task and 51.1\% MS mIoU on ADE20K \cite{zhou2017scene} semantic segmentation task.

\begin{figure*}[t]
    \centering
    \vspace{-1em}
    \includegraphics[width=\linewidth]{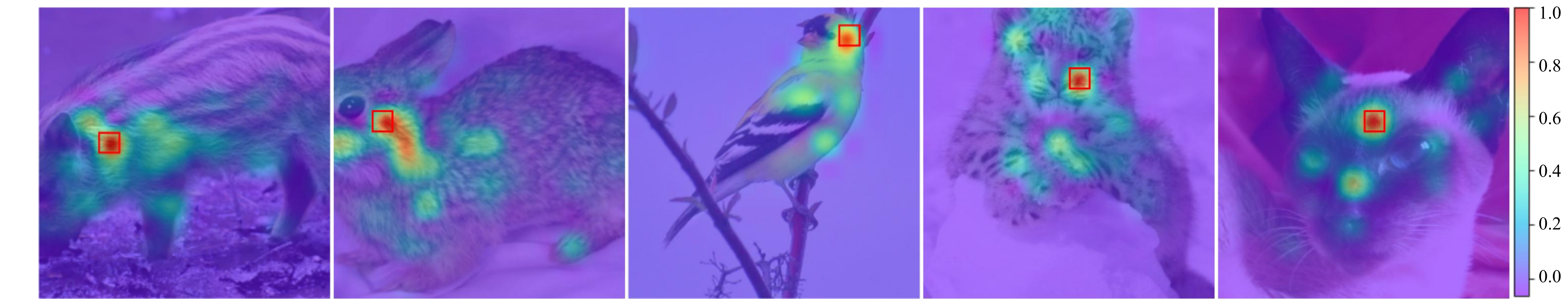}
    \vspace{-1.5em}
    \caption{Visualization of attention maps of the third Multi-Head Self-Attention block of ViT-Small\protect\footnotemark[1]. We visualize the activations in attention maps of the query patches (in the red box).
   The attention maps show that patches with high attention scores sparsely scatter around the query patch, and other patches have low attention scores.
   }
    \vspace{-1.5em}
    \label{atten_fig}
\end{figure*}

\stepcounter{footnote}\footnotetext{We use the official checkpoint from \url{https://github.com/google-research/vision_transformer}}

\section{Related Work}
\label{related_work}
A comparison of technical details with various models is shown in Table \ref{tab:other_model}. We summarize and classify our DilateFormer and related vision transformer models from the perspectives of overlapping tokenizer/downsampler, positional embedding, attention type and multi-scale. In the following section, we detail some related works.

\begin{table*}[t]
\small
\vspace{-1em}
\begin{center}
\captionsetup{justification=centering}
\caption{\textsc{Comparison of technical details with other models. ``-'' indicates these modules do not exist. For overlapping tokenizer/downsampler, ``$\checkmark$'' and ``×'' indicate whether these modules are overlapping or not.  For positional embedding, ``APE'', ``RPE'' and ``CPE'' indicate absolute positional embedding, relative positional embedding and convolutional positional embedding, respectively.  For other technical details, ``$\checkmark$'' and ``×'' indicate these modules are used or not.}} 
\label{tab:other_model}
\vspace{-0.5em}
\resizebox{\textwidth}{!}{
\begin{tabular}{ccccccccc}
\toprule
\multirow{2}{*}{Model} & Overlapping   & Overlapping       & Positional & \multicolumn{3}{c}{Attention}    & \multicolumn{2}{c}{Multi-scale}           \\
                       & Tokenizer     & Downsampler       & Embedding  &  Local    & Global    & Sparse   & Stage-level     & Block-level      \\
\midrule
ViT\cite{dosovitskiy2020image}/DeiT\cite{deit} & × & - & APE & × & $\checkmark$ & × & ×  & ×  \\
PVT\cite{wang2021pyramid}               & × & × & APE & × & $\checkmark$ & × & $\checkmark$   & ×   \\
Swin\cite{liu2021swin}                  & × & × & RPE & $\checkmark$ & × & × & $\checkmark$   & ×   \\
Twins\cite{chu2021twins}                & × & × & CPE & $\checkmark$ & $\checkmark$ & × & $\checkmark$  & ×    \\
GG\cite{yu2021glance}                   & × & × & RPE/APE &  × & × &  $\checkmark$  & $\checkmark$   & ×    \\
Shuffle\cite{Huang2021shuffle}          & $\checkmark$  & × & RPE       & $\checkmark$ & × & $\checkmark$ & $\checkmark$   & ×   \\
MaxViT\cite{tu2022maxvit}               & $\checkmark$  & $\checkmark$  & CPE & $\checkmark$ & × & $\checkmark$   & $\checkmark$   & ×   \\
CrossFormer\cite{Wang2021crossf}        & $\checkmark$  & $\checkmark$  & RPE & $\checkmark$ & × & $\checkmark$   & $\checkmark$  & $\checkmark$  \\
ViL\cite{zhang2021vil}                  & × & × & RPE/APE & $\checkmark$ & $\checkmark$ & ×   & $\checkmark$   & ×  \\
NAT\cite{Hassani2022nat}                & $\checkmark$  & $\checkmark$  & RPE  & $\checkmark$ & × & × & $\checkmark$   & ×   \\
Mobile-Former\cite{chen2021Mobileformer} & - & - & CPE & × & $\checkmark$ & × & $\checkmark$  & ×    \\
Conformer\cite{peng22021conformer}       & $\checkmark$  & -  & -  & × & $\checkmark$ &  ×   & $\checkmark$   & ×   \\
Shunted\cite{ren2021Shunted}             & $\checkmark$  & $\checkmark$  & CPE  & × & $\checkmark$  & ×  & $\checkmark$   & $\checkmark$     \\
MPViT\cite{ren2021Shunted}               & $\checkmark$  & $\checkmark$  & CPE  & × & $\checkmark$  & × & $\checkmark$   & $\checkmark$    \\
ViTAE\cite{xu2021vitae}                  & $\checkmark$  & -             & APE  & × & $\checkmark$  & × & $\checkmark$   & $\checkmark$    \\
UniFormer\cite{li2022uniformer}          & $\checkmark$  & $\checkmark$  & CPE  & × & $\checkmark$ & ×  & $\checkmark$   & ×   \\
Focal\cite{yang2021focal}                & × & × & RPE        & $\checkmark$ & $\checkmark$ & × & $\checkmark$   & $\checkmark$  \\
\rowcolor{gray!20} DilateFormer (ours)   & $\checkmark$  & $\checkmark$  & CPE &  $\checkmark$ & $\checkmark$   & $\checkmark$  & $\checkmark$   & $\checkmark$    \\
\bottomrule
\end{tabular}
}
\end{center}
\vspace{-2em}
\end{table*}

\subsection{Global Attention in Vision Transformers}
\label{global_atten}

Inspired by the success in NLP \cite{Devlin2019bert,Vaswani2017atten,Brown2020Language}, the vanilla Vision Transformers (ViTs) \cite{dosovitskiy2020image,deit} directly apply self-attention mechanisms to patches split from images. By utilizing sufficient training data  \cite{zhai2021scale,dosovitskiy2020image} and strong data augmentation strategies \cite{deit,zhang2018mixup,Yun2019cutmix,Szegedy2016smooth,Hoffer2020Repetition,Zhong2020Erasing}, Transformer-based methods \cite{wu2021rpe,yue2021psvit,yuan2021volo,zhang2022nested,pisltrc2022pan,ZhangZCHLJ22,LinYXYZL22,MaGSMKCTYF21} achieve exciting performance improvements on various vision tasks, \ie, image classification \cite{deit,liu2021swin,li2022uniformer,li2021ContextualTrans,dong2021cswin,peng22021conformer}, object detection \cite{zhang2021vil,dong2021cswin,Hassani2022nat,Carion2020tettrans,gao2021fastdetr,dai2021dynamichead,MaGSMKCTYF21,LiPLW21}, semantic segmentation \cite{ren2021Shunted,lee2021mpvit,chu2021twins,guo2021sotr,zhu2021pvtss,LinYXYZL22,ZhouGLF21}, and re-identification \cite{chen2022rest,he2021transreid,zheng2022template}. Since the computational complexity of the self-attention mechanism is quadratic \textit{w.r.t.} the number of patches, global attention is difficult to apply in high-resolution image encoding. Furthermore, according to our analysis in Sec. \ref{intro}, the long-range modeling capability of the global attention mechanism in shallow layers of ViTs is redundant. To reduce the redundancy and computational cost of the self-attention mechanism, some works \cite{wang2021pyramid, chu2021twins, yu2021glance} introduce sub-sampling operations in self-attention blocks while preserving the global receptive field. 
Such sub-sampling operations require complex designs and introduce extra parameters or computational cost. 
Different from these works, our Sliding Window Dilated Attention (SWDA) is  easy to implement for reducing the redundancy of self-attention mechanism in a dilated manner.

\subsection{Local Attention in Vision Transformers}
\label{local_atten}
In order to make the self-attention mechanism applicable for high-resolution image encoding, some works \cite{liu2021swin,dong2021cswin,GongYLFLFL22} apply the self-attention mechanism to patches in a fixed local region to reduce computational cost. 
For example, Swin \cite{liu2021swin} applies self-attention to the patches within fixed windows 
and then adopts a window-shifting strategy in the next layer for information exchange between the patches in different windows. 
CSwin \cite{dong2021cswin} improves the window-fixed setting in Swin \cite{liu2021swin}, performing self-attention to cross-shaped windows. 
Other works \cite{Huang2021shuffle,Wang2021crossf, tu2022maxvit} use grouped sampling or spatial shuffling operation for information exchange between different local windows.
Inspired by the convolution operation in CNNs \cite{AlexNet,resnet,googlenet,liu2022ConvNeXt,li2021simvit}, ViL \cite{zhang2021vil} and NAT \cite{Hassani2022nat} propose sliding window attention, which models dependencies only with neighboring patches in the window centering each query patch. Moreover, some works  \cite{xiao2021EarlyConv,peng22021conformer,chen2021Mobileformer,guo2021cmt,yuan2021ceit,Srinivas2021BoTNet,XiaLL22} combine CNNs and Transformers for introducing the locality prior, and they usually design hand-crafted and complex modules for interaction between CNNs and Transformers features, leading to a lack of scalability to large-scale parameters \cite{kaiming2021mae,xie2021simim}.  However, some works \cite{liu2021swin, dong2021cswin,Hassani2022nat,zhang2021vil} above only consider the locality of the self-attention mechanism and lack consideration of the sparsity. Although some works \cite{yu2021glance, Huang2021shuffle, tu2022maxvit, Wang2021crossf} above perform self-attention in a sparse and uniform manner, 
they are designed to approximate the global attended receptive field. 
In comparison, our Sliding Window Dilated Attention (SWDA) takes both the locality and sparsity of self-attention mechanism into consideration. Our SWDA introduces a prior to reduce the redundancy of self-attention mechanism, which performs self-attention in a dilated window centered on query patch.

\subsection{Multi-scale Vision Transformer}
\label{multi_scale}
The vanilla Vision Transformer \cite{dosovitskiy2020image, deit} is a “columnar” structure for visual tasks. Since multi-scale information \cite{AlexNet,resnet,googlenet,YuZW22,LiuLLSL22,ChenLLM22,LiuFWLSZ22,ZuoWFHSW22} is beneficial for dense prediction tasks such as object detection, instance and semantic segmentation, recent works \cite{wang2021pyramid,liu2021swin,dong2021cswin, zhang2021vil,Hassani2022nat, ren2021Shunted,Huang2021shuffle,yu2022metaformer,lee2021mpvit,Wang2021crossf,fan2021multiscale,MaGSMKCTYF21} introduce multi-scale modeling capability by using a pyramid structure to design their transformer backbones. 
Several works \cite{Wang2021crossf,chen2021crossvit,lee2021mpvit,ren2021Shunted, xu2021vitae,yang2021focal,peng22021conformer,chen2021Mobileformer} introduce multi-scale information in patch embedding layers \cite{Wang2021crossf} or self-attention blocks \cite{lee2021mpvit,ren2021Shunted,yang2021focal} or add extra branches \cite{peng22021conformer,chen2021Mobileformer,xu2021vitae} to perform convolution operation. CrossFormer\cite{Wang2021crossf}  utilizes different convolution operations or different patch sizes for designing patch embedding. Shunted Transformer\cite{ren2021Shunted} uses multi-scale token aggregation for obtaining keys and values of various sizes. MPViT \cite{lee2021mpvit} consists of multi-scale patch embedding and multi-path transformer blocks. Conformer \cite{peng22021conformer},  Mobile-Former \cite{chen2021Mobileformer} and ViTAE \cite{xu2021vitae} 
design additional convolution branches outside or inside the self-attention blocks to integrate multi-scale information. The above methods all require complex design, which inevitably introduce additional parameters and computational cost. Our Multi-Scale Dilated Attention (MSDA) extracts multi-scale features by setting different dilation rates, which is simple and does not need to introduce extra parameters and computational cost.

\subsection{Dilated Convolution}
\label{dilate_conv2d}
Traditional Convolution-based networks \cite{AlexNet,resnet,googlenet,liu2022ConvNeXt} usually use downsampling or convolution with a large stride to increase the receptive field and reduce computational cost. However, these approaches \cite{AlexNet,resnet,googlenet,liu2022ConvNeXt} result in reduced resolution of feature maps, affecting model performance in many tasks such as object detection \cite{fastrcnn,yolov3,ssd,retinanet,he2017mask} and semantic segmentation \cite{unet,DeepLabV3,fcn}. Therefore, Cohen et al. \cite{yu2015multi,cohen2016group} propose dilated convolution\cite{MaSC22,YanZZZZ22}, which increases the receptive field without reducing the resolution and extracts the information of the feature map at different scales by setting different dilation rates.  Dilated convolution with dynamic weights \cite{chen2020dynamic}, namely Dynamic Dilated Convolution (DDC), uses the entire feature map to generate the kernel parameter of convolution, which is data-specific at the feature-map level.

Different from existing works, we propose a simple yet effective Dilated Attention operation by introducing various dilation rates at the same semantic level into a single self-attention operation, which more flexibly models multi-scale interaction. Although ours is a sliding window based dilated attention, ours differs from DDC because our modelling performs self-attention on keys and values sparsely selected in a sliding window centered on the query patch, which is data-specific at the token level. In addition, we also notice a concurrent work, DiNAT \cite{hassani2022dilated}, which uses a single-scale and fixed dilation rate in each block of the same stage, lacking multi-scale interaction. In contrast, our DilateFormer uses a multi-scale strategy in each block i.e., setting different dilation rates for
different heads, which can capture and fuse multi-scale semantic feature.

\section{Multi-Scale Dilated Transformer}
\label{method}
In this section, we introduce our proposed Multi-Scale Dilated Transformer (DilateFormer) in details. In Sec.\ref{dilate_atten}, we introduce our Sliding Window Dilated Attention (SWDA) operation, towards effective long-range dependency modeling in feature maps. In Sec.\ref{dilate_block}, we design Multi-Scale Dilated Attention (MSDA), which simultaneously captures contextual semantic dependencies at different scales to make good use of the information inside the block. The overall framework and variants of the proposed Multi-Scale Dilated Transformer (DilateFormer) are illustrated in Sec.\ref{Arc}.

\begin{figure*}[t]
    \centering
    \vspace{-1em}
    \includegraphics[width=\textwidth]{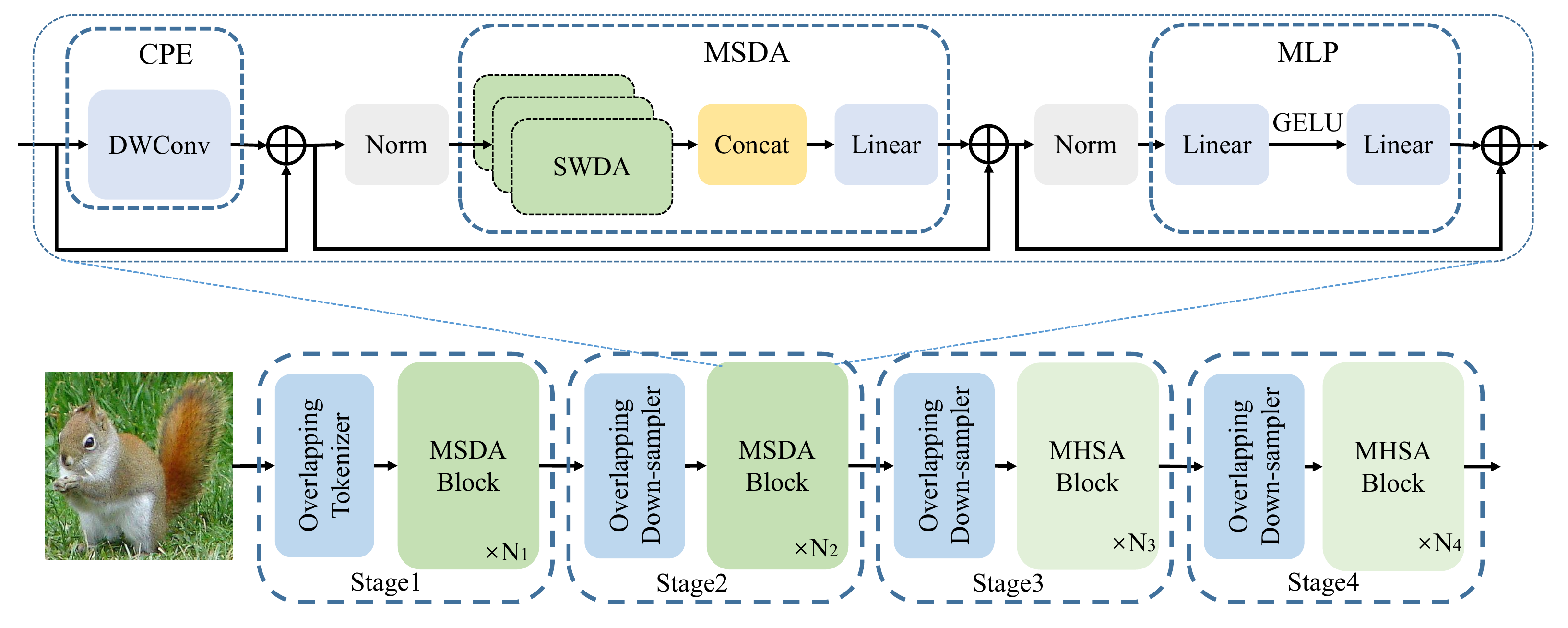}
    \vspace{-2em}
    \caption{The overall architecture of our DilateFormer. The top part shows the proposed Multi-Scale Dilated Attention (MSDA) block, consisting of DwConv, Multi-Scale Sliding Window Dilated Attention operation (SWDA) and MLP. The bottom part shows DilateFormer, consisting of Overlapping Tokenizer, Overlapping Downsampler, Multi-Scale Dilated Attention (MSDA) block and Multi-Head Self-Attention (MHSA) block.}
    \label{arc_fig}
    \vspace{-1em}
\end{figure*}

\begin{figure}[t]
    \centering
    \vspace{-1em}
    \includegraphics[width=\linewidth]{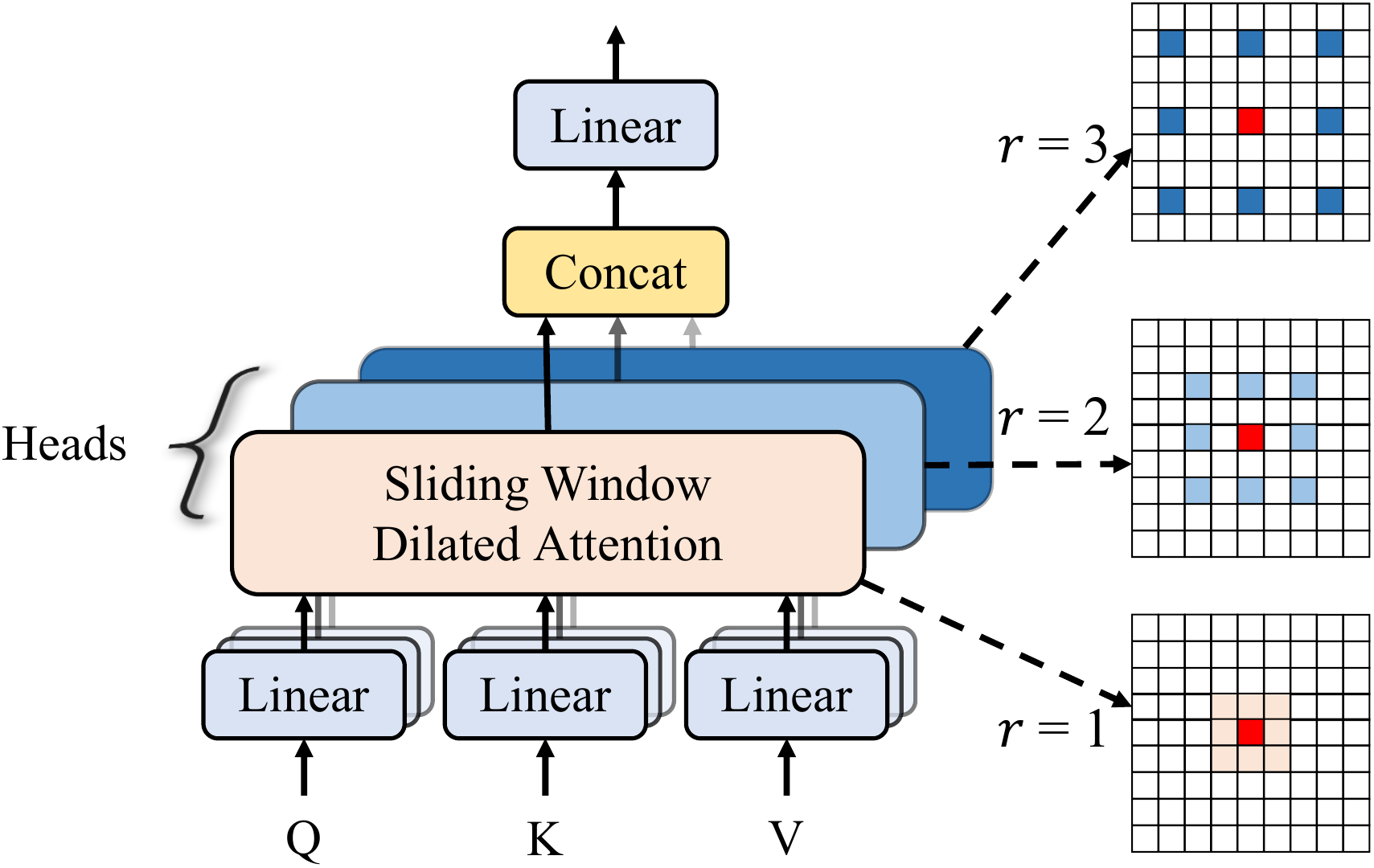}
    \caption{\textbf{Illustration of Multi-Scale Dilated Attention (MSDA).} 
    First, the channels of the feature map are split into different heads. Then, the self-attention operation is performed among the colored patches in the window surrounding the red query patch, using different dilation rates in different heads.
    Besides, features in different heads are concatenated together and then fed into a linear layer.
    By default, we use a $3\times3$ kernel size with dilation rates $r$ = 1, 2 and 3, and the sizes of attended receptive fields in different heads are $3\times3$, $5\times5$ and $7\times7$.}
        \vspace{-1em}
    \label{msda_fig}
\end{figure}

\subsection{Sliding Window Dilated Attention}
\label{dilate_atten}
According to the locality and sparsity properties observed in the global attention of shallow layers in vanilla Vision Transformers (ViTs), we propose a Sliding Window Dilated Attention (SWDA) operation, where the keys and values are \textit{sparsely} selected in a sliding window centered on the query patch. Self-attention is then performed on these representative patches. Formally, our SWDA is described as follows:
\begin{equation}
\label{swda}
\vspace{-0.5em}
    X = \mathrm{SWDA}(Q,K,V,r), 
\end{equation}
where $Q$, $K$ and $V$ represent the query, key and value matrix, respectively. Each row of the three matrices indicates a single query/key/value feature vector. For the query at location $(i,~j)$ in the original feature map,
SWDA sparsely selects keys and values to conduct self-attention in a sliding window of size $w\times w$ centered on $(i,~j)$. Furthermore, we define a dilation rate $r\in\mathbb{N}^+$ to control the degree of sparsity. Particularly, for the position $(i,~j)$, the corresponding component $x_{ij}$ of the output $X$ from $\mathrm{SWDA}$ operation is defined as follows:
\begin{equation}
\begin{aligned}
\label{softmax}
    x_{ij} & = \mathrm{Attention}(q_{ij}, K_{r}, V_{r}),\\
    & = \mathrm{Softmax}\left(\frac{q_{ij} K^{T}_{r}}{\sqrt{d_{k}}} \right)V_{r}, ~~1 \leq i \leq W, ~1 \leq j \leq H,
\end{aligned}
\end{equation}
where $H$ and $W$ are the height and width of the feature map.
$K_{r}$ and $V_{r}$ represent keys and values selected from the feature maps $K$ and $V$. 
Given the query positioned at $(i,~j)$, keys and values positioned at the following set of coordinate $(i',~j')$ will be selected for conducting self-attention:
\begin{equation}
\begin{aligned}
\label{index}
   \Big\{(i',j') \Big|& 
    i'=i+p \times r, 
    j'=j+q \times  r  
   \Big\}, 
   \\ & -\frac{w}{2} \leq p,~q \leq \frac{w}{2}.
\end{aligned}
\end{equation}

Our SWDA conducts the self-attention operation for all query patches in a sliding window manner. For the query at the edge of the feature map, we simply use the zero padding strategy commonly used in convolution operations to maintain the size of the feature map. By sparsely selecting keys and values centered on queries, the proposed SWDA explicitly satisfies the locality and sparsity property and can model the long-range dependency effectively.

\subsection{Multi-Scale Dilated Attention}
\label{dilate_block}

To exploit the sparsity at different scales of the self-attention mechanism in block-level, we further propose a Multi-Scale Dilated Attention (MSDA) block to extract multi-scale semantic information. As shown in Figure \ref{msda_fig}, given a feature map $X$, we obtain corresponding queries, keys and values by linear projection.
After that, we divide the channels of the feature map to $n$ different heads and perform multi-scale SWDA in different heads with different dilation rates.
Specifically, our MSDA is formulated as follows:
\begin{equation}
    h_{i} = \mathrm{SWDA}(Q_{i},K_{i},V_{i},r_{i}),~~~~1 \leq i \leq n,
\end{equation}
\begin{equation}
    X = \mathrm{Linear}(\mathrm{Concat}[h_{1},...,h_{n}]),
\end{equation}
where $r_{i}$ is the dilation rate of the $i$-th head and $Q_{i}$, $K_{i}$ and $V_{i}$ represent slices of feature maps fed into the $i$-th head. The outputs $\{h_i\}_{i=1}^n$ are concatenated together and then sent to a linear layer for feature aggregation. 

By setting different dilation rates for different heads, our MSDA effectively aggregates semantic information at various scales within the attended receptive field and efficiently reduces the redundancy of self-attention mechanism without complex operations and extra computational cost.

\begin{table*}[t]
\centering
\small
\vspace{-1em}
\captionsetup{justification=centering}
\caption{\textsc{Model variants of our DilateFormer. MSDA and MHSA represent Multi-Scale Dilated Attention and Multi-Head Self-Attention, respectively.} ``d'', ``h'', ``ks.'' \textsc{and}  ``dr.''\textsc{ indicate feature dimension, the number of head, kernel size and dilation rate, respectively.}}
 \vspace{-0.5em}
\setlength{\tabcolsep}{6pt} 
\renewcommand{\arraystretch}{1.5} 
\begin{tabular}{ccccc}
\hline
Resolution & Block & Tiny  & Small & Base \\
\hline
\begin{tabular}[c]{@{}c@{}@{}}Stage 1\\ $(56\times56)$\\\end{tabular}&  {\begin{tabular}[c]{@{}c@{}}MSDA\end{tabular} } & \begin{tabular}[c]{@{}c@{}}\specialrule{0em}{1pt}{3pt}$\begin{bmatrix}72\text{-d, } 3\text{-h}\\\text{ks. }3\times3 \\\text{dr. } [1,2,3]\end{bmatrix}\times 2$\\ \end{tabular}  & 
\begin{tabular}[c]{@{}c@{}}\specialrule{0em}{1pt}{3pt}$\begin{bmatrix}72\text{-d, } 3\text{-h}\\\text{ks. }3\times3 \\\text{dr. } [1,2,3]\end{bmatrix}\times 3$\\ \end{tabular}  & 
\begin{tabular}[c]{@{}c@{}}\specialrule{0em}{1pt}{3pt}$\begin{bmatrix}96\text{-d, } 3\text{-h}\\\text{ks. }3\times3 \\\text{dr. } [1,2,3]\end{bmatrix}\times 4$\\ \end{tabular}  \\
\hline
\begin{tabular}[c]{@{}c@{}@{}}Stage 2\\ $(28\times28)$\\\end{tabular}&  {\begin{tabular}[c]{@{}c@{}}MSDA\end{tabular} } & \begin{tabular}[c]{@{}c@{}}\specialrule{0em}{1pt}{3pt}$\begin{bmatrix}144\text{-d, } 6\text{-h}\\\text{ks. }3\times3 \\\text{dr. } [1,2,3]\end{bmatrix}\times 2$\\ \end{tabular}  & 
\begin{tabular}[c]{@{}c@{}}\specialrule{0em}{1pt}{3pt}$\begin{bmatrix}144\text{-d, } 6\text{-h}\\\text{ks. }3\times3 \\\text{dr. } [1,2,3]\end{bmatrix}\times 5$\\ \end{tabular}  & 
\begin{tabular}[c]{@{}c@{}}\specialrule{0em}{1pt}{3pt}$\begin{bmatrix}192\text{-d, } 6\text{-h}\\\text{ks. }3\times3 \\\text{dr. } [1,2,3]\end{bmatrix}\times 8$\\ \end{tabular}  \\
\hline
\begin{tabular}[c]{@{}c@{}@{}}Stage 3\\ $(14\times14)$\\\end{tabular}&  {\begin{tabular}[c]{@{}c@{}}MHSA\end{tabular} } & \begin{tabular}[c]{@{}c@{}}\specialrule{0em}{1pt}{3pt}$\begin{bmatrix}288\text{-d, } 12\text{-h} \end{bmatrix}\times 6$\\ \end{tabular}  & 
\begin{tabular}[c]{@{}c@{}}\specialrule{0em}{1pt}{3pt}$\begin{bmatrix}288\text{-d, } 12\text{-h} \end{bmatrix}\times 8$\\ \end{tabular}  & 
\begin{tabular}[c]{@{}c@{}}\specialrule{0em}{1pt}{3pt}$\begin{bmatrix}384\text{-d, } 12\text{-h} \end{bmatrix}\times 10$\\\end{tabular}  \\
\hline
\begin{tabular}[c]{@{}c@{}@{}}Stage 4\\ $(7\times7)$\\\end{tabular}&  {\begin{tabular}[c]{@{}c@{}}MHSA\end{tabular} } & \begin{tabular}[c]{@{}c@{}}\specialrule{0em}{1pt}{3pt}$\begin{bmatrix}576\text{-d, } 24\text{-h}\end{bmatrix}\times 2$\\ \end{tabular}  & 
\begin{tabular}[c]{@{}c@{}}\specialrule{0em}{1pt}{3pt}$\begin{bmatrix}576\text{-d, } 24\text{-h}\end{bmatrix}\times 3$\\ \end{tabular}  & 
\begin{tabular}[c]{@{}c@{}}\specialrule{0em}{1pt}{3pt}$\begin{bmatrix}768\text{-d, } 24\text{-h}\end{bmatrix}\times 3$\\ \end{tabular}  \\
\hline
\end{tabular}
\normalsize
\label{Model_table}
\vspace{-1em}
\end{table*}

\subsection{Overall Architecture}
\label{Arc}

With a pyramid structure, we propose the Multi-Scale Dilated Transformer (DilateFormer) as shown in Figure \ref{arc_fig}.
According to the locality and sparsity property of shallow layers in ViTs, the first two stages of DilateFormer use Multi-Scale Dilated Attention (MSDA) proposed in Sec. \ref{dilate_block} while the latter two stages utilize ordinary Multi-Head Self-Attention (MHSA).
What's more, we use the overlapping tokenizer \cite{xiao2021EarlyConv} for patch embedding, which uses multiple overlapping $3\times 3$ convolution modules with zero-padding. The resolution of the output feature map can be adjusted by controlling the stride size of convolution kernels to be 1 or 2 alternately. 
To merge patches in the previous stage, we utilize the overlapping downsampler \cite{Hassani2022nat}, a convolution module with an overlapping kernel size of 3 and a stride of 2.
To make the position encoding adaptive to inputs of different resolutions, we use Conditional Position Embedding (CPE) proposed in CPVT \cite{chu2021cpe} whenever inputs are fed into MSDA or MHSA blocks. Specifically, our overall architecture is described as follows:
\begin{equation}
    X = \mathrm{CPE}(\hat{X}) + \hat{X} = \mathrm{DwConv}(\hat{X}) + \hat{X},  
\end{equation}
\begin{equation}
Y =
\begin{dcases}
  \mathrm{MSDA}(\mathrm{Norm}(X)) + X, & \text{at~low-level~stages}, \\
  \mathrm{MHSA}(\mathrm{Norm}(X)) + X, & \text{at~high-level~stages},
\end{dcases}
\end{equation}
\begin{equation}
    Z = \mathrm{MLP}(\mathrm{Norm}(Y)) + Y,
\end{equation}
where $\hat{X}$ is the input of the current block, \ie, the image patches or the output from the last block. In practice, we implement CPE as a depth-wise convolution (DwConv) module with zero-padding and $3\times 3$ kernel size. 
We add MLP following prior works \cite{deit,liu2021swin}, which consists of two linear layers with the channel expansion ratio of 4 and one GELU activation.

Based on the above network structure, we introduce three variants of the proposed DilateFormer (\ie, Tiny, Small, and Base), and the specific model settings are given in Table \ref{Model_table}.

\section{Experiments}
\label{exp}
To evaluate the performance of our Multi-Scale Dilated Transformer (DilateFormer), we take our model as a vision backbone for ImageNet-1K \cite{deng2009large} classification, COCO \cite{lin2014microsoft} object detection and instance segmentation, and ADE20K \cite{zhou2017scene} semantic segmentation. Furthermore, we evaluate the effectiveness of our key modules via ablation studies. All experiments are conducted on a single server node with 8 A100 GPUs.

\begin{table}
    \centering
    \vspace{-1em}
    \captionsetup{justification=centering}
    \caption{\textsc{Comparison with the state-of-the-art on ImageNet-1K. `$\star$' indicates Token Labeling proposed in LV-ViT \cite{jiang2021tlt}, and `$\uparrow$' indicates that the model is fine-tuned at a larger resolutions.}}
    \vspace{-0.5em}
    \resizebox{\linewidth}{!}{
    \begin{tabular}{c|cc|cc|c}
    \hline
    \multirow{2}{*}{Method} & Params  & FLOPs  & \multirow{2}{*}{Train} & \multirow{2}{*}{Test} & Top1 \\
     &  (M) & (G) & &  & (\%) \\
    \hline
    \hline
    RegNetY-4G \cite{radosavovic2020designing}   & 21 & 4.0 & 224 & 224 & 80.0 \\
    ResNet-50 \cite{resnet} & 25 & 4.1 & 224 & 224 & 78.5 \\
    ConvNeXt-T \cite{liu2022ConvNeXt}   & 28 & 4.5 & 224 & 224 & 82.1 \\
    Mobile-Former-508M \cite{chen2021Mobileformer} &	14 &	1.0	 & 224  & 224 &  79.3 \\
    PVT-S \cite{wang2021pyramid}         & 25 & 3.8 & 224 & 224 & 79.8 \\
    DW-Conv.-T \cite{han2021connection}  & 24 & 3.8 & 224 & 224 & 81.3 \\
    CoAtNet-0 \cite{dai2021coatnet}    & 25 & 4.2 & 224 & 224 & 81.6 \\
    Swin-T \cite{liu2021swin}        & 29 & 4.5 & 224 & 224 & 81.3 \\
    CvT-13 \cite{wu2021cvt}    & 20 & 4.5 & 224 & 224 & 81.6 \\
    GG-T \cite{yu2021glance}    & 28 & 4.5 & 224 & 224 & 82.0 \\
    DeiT-S \cite{deit}        & 22 & 4.6 & 224 & 224 & 79.9 \\
    Distilled DeiT-S \cite{deit}        & 22 & 4.6 & 224 & 224 & 81.2 \\
    ViL-S \cite{zhang2021vil}         & 25 & 4.9 & 224 & 224 & 82.0 \\
    TNT-S \cite{han2021tnt} & 24 & 5.2 & 224 & 224 & 81.3 \\
    NesT-T \cite{zhang2021aggregating}       & 17 & 5.8 & 224 & 224 & 81.3 \\
    BoTNet-S1-59 \cite{Srinivas2021BoTNet}   & 34 & 7.3 & 224 & 224 & 81.7 \\
    \rowcolor{gray!20} Dilate-T  (ours)          & 17 & 3.2 & 224 & 224 & 82.1 \\ 
    \rowcolor{gray!20} Dilate-T$^{\star}$  (ours)   & 18 & 3.2 & 224 & 224 & \textbf{82.8} \\
    \midrule
    CvT-13 $\uparrow384$ \cite{wu2021cvt} & 20 & 16.3 & 224 & 384 & 83.0 \\
    \rowcolor{gray!20} Dilate-T$^{\star}\uparrow384$  (ours) & 18 & 10.2 & 224 & 384 & \textbf{83.8}\\
    \hline
    \hline
    ResNet-101 \cite{resnet}    & 44 & 7.9 & 224 & 224 & 79.8 \\
    ConvNeXt-S \cite{liu2022ConvNeXt}  & 50 & 8.7 & 224 & 224 & 83.1  \\
    RegNetY-16G \cite{radosavovic2020designing}   & 84 & 16.0 & 224 & 224 & 82.9 \\
    Focal-T \cite{yang2021focal}       & 29 & 4.9 & 224 & 224 & 82.2 \\
    CrossFormer-S \citeonline{Wang2021crossf} & 31 & 4.9 & 224 & 224 & 82.5 \\
    T2T-14 \cite{yuan2021tokens}        & 22 & 5.2 & 224 & 224 & 80.7  \\
    DiNAT-T \cite{hassani2022dilated} &	28  & 4.3 & 224  & 224 &	 82.7 \\
    LV-ViT-S$^{\star}$ \cite{jiang2021tlt}  & 26 & 6.6 & 224 & 224 & 83.3 \\
    CvT-21 \cite{wu2021cvt}         & 32 & 7.1 & 224 & 224 & 82.5 \\
    Twins-SVT-B \cite{chu2021twins} & 56 & 8.3 & 224 & 224 & 83.1 \\ 
    Swin-S \cite{liu2021swin}       & 50 & 8.7 & 224 & 224 & 83.0  \\
    PoolFormer-M36 \cite{yu2022metaformer} & 56 & 8.8 & 224 & 224 & 82.1 \\
    PVT-L \cite{wang2021pyramid}    & 61 & 9.8 & 224 & 224 & 81.7 \\
    NesT-S \cite{zhang2021aggregating}   & 38 & 10.4 & 224 & 224 & 83.3 \\
    DeepVit-L          & 55 & 12.5 & 224 & 224 & 82.2\\
    CoaT-S         & 22 & 12.6 & 224 & 224 & 82.1 \\
    TNT-B \cite{han2021tnt} &  66 &  14.1 & 224 & 224 &  82.8 \\
    \rowcolor{gray!20} Dilate-S  (ours)          & 21 & 4.8 & 224 & 224 & 83.3 \\
    \rowcolor{gray!20} Dilate-S$^{\star}$  (ours)  & 22 & 4.9 & 224 & 224 & \textbf{83.9} \\
    \hline
    CoAtNet-0$\uparrow384$ \cite{dai2021coatnet}  & 20 & 13.4 & 224 & 384 & 83.9 \\
    T2T-14 $\uparrow384$ \cite{yuan2021tokens}  & 22  & 17.1 & 224 & 384 & 83.3 \\
    LV-ViT-S$^{\star} \uparrow384$ \cite{jiang2021tlt} & 26 & 22.2 & 224 & 384 & 84.4 \\
    CvT-21 $\uparrow384$ \cite{wu2021cvt} & 32 & 24.9 & 224 & 384 & 83.3\\
    \rowcolor{gray!20} Dilate-S$^{\star}\uparrow384$  (ours) & 22 & 15.5 & 224 & 384 & \textbf{84.9} \\
    \hline
    \hline
    ResNet-152 \cite{resnet}    & 60 & 11.6 & 224 & 224 & 80.8 \\
    EffNet-B7 \cite{tan2019efficientnet}  & 54 & 39.2 & 600 & 600 & 84.3  \\
    Next-ViT-L \cite{li2022next}  & 58 & 10.8 & 224 & 224 & 83.6 \\
    PoolFormer-M48 \cite{yu2022metaformer} & 73 & 11.6 & 224 & 224 & 82.5 \\
    DeepViT-L \cite{zhou2021DeepViT} & 55 & 12.5 & 224 & 224 & 83.1 \\
    DW-Conv.-B \cite{han2021connection} & 74 & 12.9 & 224 & 224 & 83.2 \\
    DiNAT-S \cite{hassani2022dilated} &	51	& 7.8 &  224  & 224  &	83.8 \\
    T2T-24  \cite{yuan2021tokens}    & 64 & 13.2 & 224 & 224 & 82.2 \\
    ViL-B  \cite{zhang2021vil}       & 56 & 13.4 & 224 & 224 & 83.2 \\
    Twins-SVT-L \cite{chu2021twins} & 99 & 14.8 & 224 & 224 & 83.3 \\ 
    Swin-B \cite{liu2021swin}        & 88 & 15.4 & 224 & 224 & 83.4  \\
    Shuffle-B \cite{Huang2021shuffle} & 88 & 15.6 & 224 & 224 & 84.0 \\
    CoAtNet-2 \cite{dai2021coatnet}  & 75 & 15.7 & 224 & 224 & 84.1 \\
    Focal-B \cite{yang2021focal}      & 90 & 16.0 & 224 & 224 & 83.8 \\
    LV-ViT-M$^{\star}$ \cite{jiang2021tlt}  & 56 & 16.0 & 224 & 224 & 84.1\\
    CrossFormer-L \cite{Wang2021crossf} & 92 & 16.1 & 224 & 224 & 84.0 \\
    MPViT-B \cite{lee2021mpvit}       & 75 & 16.4 & 224 & 224 & 84.3 \\
    DeiT-B \cite{deit}       & 86 & 17.5 & 224 & 224 & 83.4 \\
    Distilled DeiT-B \cite{deit}       & 86 & 17.5 & 224 & 224 & 81.8 \\
    NesT-B  \cite{zhang2021aggregating}       & 68 & 17.9 & 224 & 224 & 83.8 \\
    BoTNet-T7 \cite{Srinivas2021BoTNet}          & 79 & 19.3 & 256 & 256 & 84.2 \\
    \rowcolor{gray!20} Dilate-B   (ours)         & 47 & 10.0 & 224 & 224 & 84.4\\ 
    \rowcolor{gray!20} Dilate-B$^{\star}$  (ours)  & 48 & 10.0 & 224 & 224 & \textbf{84.9}\\
    \hline
    CoAtNet-1 $\uparrow384$ \cite{dai2021coatnet}  & 42 & 27.4 & 224 & 384 & 85.1 \\
    LV-ViT-M$^{\star}\uparrow384$ \cite{jiang2021tlt} & 56 & 42.2 & 224 & 384 & 85.4\\
    BoTNet-S1-128$\uparrow384$ \cite{Srinivas2021BoTNet}  & 79 & 45.8 & 256 & 384 & 84.7 \\
    \rowcolor{gray!20} Dilate-B$^{\star}\uparrow384$  (ours)  & 48 & 31.1 & 224 & 384 & \textbf{85.6} \\
    \hline
    \end{tabular}
    }
    \label{cls_table} 
\end{table}

\subsection{Image Classification on ImageNet-1K}
\label{cls}
\noindent  \textbf{- Dataset and implementation details.}
ImageNet-1k \cite{deng2009large} is a large-scale 1000-classes dataset that contains 1.28 million training images and 50,000 validation images.
We conduct classification experiments on ImageNet-1K dataset to evaluate our variants, following the same training strategies of baseline Transformers as DeiT \cite{deit}  and PVT \cite{wang2021pyramid} for a fair comparison. We use the AdamW optimizer \cite{Loshchilov2019adamw} with 300 epochs including the first 10 warm-up epochs and the last 10 cool-down epochs and adopt a cosine decay learning rate scheduler decayed by a factor of 10 every 30 epochs with a base learning rate of 0.001, a batch size of 1024, and a weight decay of 0.05. To further demonstrate the performance of DilateFormer, Token Labeling  \cite{jiang2021tlt} is used to auxiliarily train DilateFormer. We add an extra fully connected layer and an auxiliary loss to DilateFormer and follow the training strategy of  LV-ViT \cite{jiang2021tlt} where CutMix \cite{zhang2018mixup} and Mixup \cite{Yun2019cutmix} are replaced by MixToken \cite{jiang2021tlt}.  For fine-tuning our models on a larger resolution, \ie, 384×384, the special hyperparameters are set as follows: weight decay, learning rate, batch size, warm-up epoch and total epoch are set to 1e-8, 5e-6, 512, 5 and 30.

\begin{figure}[t]
    \centering
    \vspace{-3em}
    \includegraphics[width=1.05\linewidth]{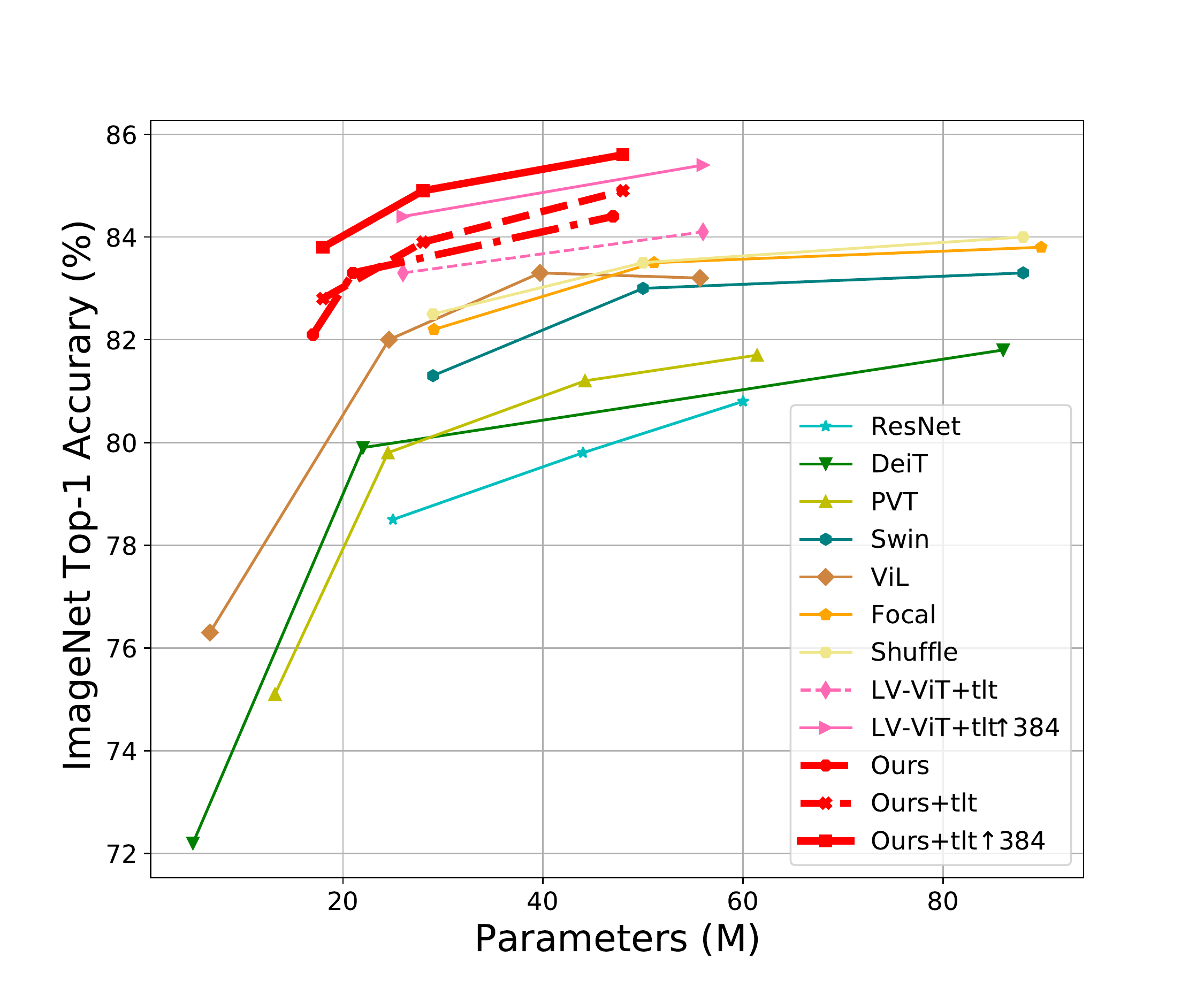}
        \vspace{-2.5em}
    \caption{Performance comparisons with respect to model parameters on ImageNet-1K classification. Without extra training data, our DilateFormer variants achieve comparable or even better performance with fewer model parameters.}
    \vspace{-1.5em}
    \label{fig:para}
\end{figure}

\noindent  \textbf{- Results and analysis.} As shown in Table \ref{cls_table}, Figure \ref{fig:flop_acc1} and Figure \ref{fig:para}, our proposed DilateFormer outperforms previous state-of-the-art models at different model sizes. 
Specifically,  Dilate-S achieves 83.3\% top-1 accuracy on ImageNet-1K with a resolution of 224, surpasses Swin-T \cite{liu2021swin}, ViL-S \cite{zhang2021vil} by 2.0\% and 1.3\% respectively and has fewer parameters and FLOPs than these models.  
With the assistance of Token Labeling \cite{jiang2021tlt} (denoted by '$\star$') , our models achieve better performance than LV-ViTs \cite{jiang2021tlt} at different model sizes, \ie, Dilate-S$^{\star}$ (4.9 GFLOPs) and Dilate-B$^{\star}$ (10.0 GFLOPs) achieve 83.9\% and 84.9\% respectively, surpassing LV-ViT-S \cite{jiang2021tlt}(6.6 GFLOPs) and LV-ViT-M \cite{jiang2021tlt} (16 GFLOPs).  
The results in Table \ref{cls_table} also show the efficiency and effectiveness of the proposed model. Without extra assistance or high-resolution finetuning, Dilate-T consumes only 3.2 GFLOPs and achieves 82.1\% accuracy, which is comparable to the performance of ViL-S \cite{zhang2021vil} (4.9G, 82.0\%), Focal-T \cite{yang2021focal} (4.9G, 82.2\%) and PVT-L \cite{wang2021pyramid}  (9.8G, 81.7\%). Similar conclusions can be found in larger models: our Dilate-S (83.3\%) with 4.8 GFLOPs outperforms ViL-B \cite{zhang2021vil} (13.4G, 83.2\%), Swin-B  \cite{liu2021swin} (15.4G, 83.4\%), and DeiT-B \cite{deit} (17.5G, 81.8\%), indicating that our MSDA can effectively capture long-range dependencies as previous methods but save up to 70\% FLOPs. 
To demonstrate the strong learning capability of DilateFormer, our Dilate-B fine-tuned on 384×384 images obtains 85.6\% top-1 accuracy and outperforms LV-ViT-M \cite{jiang2021tlt} (85.4\%) which needs 1.37 times more FLOPs.

\begin{table*}
    \centering
    \small
    \vspace{-1em}
    \captionsetup{justification=centering}
    \caption{\textsc{Object detection and instance segmentation with Mask R-CNN on COCO val2017.}}
    \vspace{-0.5em}
    \renewcommand\tabcolsep{1.5pt}
    \begin{tabular}{c|c|c|c|c|c|c|c|c|c|c|c|c|c|c}
    \toprule 
    \multirow{2}{*}{Method} 
    &\small Params & \small FLOPs & \multicolumn{6}{c|}{Mask R-CNN $1\times$ schedule} & \multicolumn{6}{c}{Mask R-CNN $3\times$ + MS schedule} \\
    & \small (M) & \small (G) & ~{AP}$^b$~ & {AP}$^{b}_{50}$ & {AP}$^{b}_{75}$ & ~{AP}$^{m}$~  & {AP}$^{m}_{50}$  & {AP}$^{m}_{75}$ & ~{AP}$^b$~ & {AP}$^{b}_{50}$ & {AP}$^{b}_{75}$ & ~{AP}$^{m}$~  & {AP}$^{m}_{50}$  & {AP}$^{m}_{75}$ \\
    \midrule
    Res50 \cite{resnet} & 44 & 260 
    & 38.0 & 58.6 & 41.4 & 34.4 & 55.1 & 36.7 
    & 41.0 & 61.7 & 44.9 & 37.1 & 58.4 & 40.1 \\
    NAT-T \cite{Hassani2022nat}    & 48 & 258
    & -    & -    & -    & -    & -     & -   
    & 47.7 & 69.0 & 52.6 & 42.6 & 66.1 & 45.9 \\
     Swin-T \cite{liu2021swin} & 48 & 264 
    & 42.2 & 64.6 & 46.2 & 39.1 & 61.6 & 42.0 
    & 46.0 & 68.2 & 50.2 & 41.6 & 65.1 & 44.8 \\
    MPViT-S \cite{lee2021mpvit} & 43 & 268
    & -    & -    & -    & -    & -     & -   
    & 48.4 & 70.5 & 52.6  & \textbf{43.9} & 67.6  & \textbf{47.5} \\
    UniFormer-S$_{h14}$ \cite{li2022uniformer} & 41 & 269 
    & 45.6 & 68.1 & 49.7 & 41.6 & 64.8 & \textbf{45.0} 
    & 48.2 & 70.4 & 52.5 & 43.4 & 67.1 & 47.0 \\
    Focal-T \cite{yang2021focal}  & 49 & 291 
    & 44.8 & 67.7 & 49.2 & 41.0 & 64.7 & 44.2
    & 47.2 & 69.4 & 51.9 & 42.7 & 66.5 & 45.9 \\
    TRT-ViT-C \cite{xia2022trt} & 86 & 294  
    & 44.7 & 66.9 & 48.8 & 40.8 & 63.9 & 44.0 
    & 47.3 & 68.8 & 51.9 & 42.7 & 65.9 & 46.0 \\
    PVT-M \cite{wang2021pyramid}  & 64 & 302 
    & 42.0 & 64.4 & 45.6 & 39.0 & 61.6 & 42.1 
    & 44.2 & 66.0 & 48.2 & 40.5 & 63.1 & 43.5 \\
    \rowcolor{gray!20} Dilate-S  (ours) & 44 &262
    & \textbf{45.8} & \textbf{68.2} &\textbf{50.1} & \textbf{41.7} & \textbf{65.3} & 44.7
    & \textbf{49.0} &  \textbf{70.9} &  \textbf{53.8} & 43.7 & \textbf{67.7} &  46.9 \\
    \midrule
    X101-32 \cite{xie2017AgRTr} &  63 &340 
    & 41.9 & 62.5 & 45.9 & 37.5 & 59.4  & 40.2
    & 44.0 & 64.4 & 48.0 & 39.2 & 61.4  & 41.9 \\
      NAT-S \cite{Hassani2022nat}   & 70 & 330 
    & -    & -    & -    & -    & -     & -   
    & 48.4 & 69.8 & 53.2 & 43.2 & 66.9 & 46.5\\
    TRT-ViT-D \cite{xia2022trt} & 121 & 375 
    & 45.3 & 67.9 & 49.6 & 41.6 & 64.7 & 44.8 
    & 48.1 & 69.3 & 52.7 & 43.4 & 66.7 & 46.8 \\
    Focal-S \cite{yang2021focal} & 71 & 401 
    & 47.4 & 69.8 & 51.9 & 42.8 & 66.6 & 46.1
    & 48.8 & 70.5 & 53.6 & 43.8 & 67.7 & 47.2 \\
    PVT-L \cite{wang2021pyramid}   & 81  & 494 
    & 42.9 & 65.0 & 46.6 & 39.5 & 61.9 & 42.5 
    & 44.5 & 66.0 & 48.3 & 40.7 & 63.4 & 43.7 \\
    Swin-B \cite{liu2021swin} & 107 & 496 
    & 46.9 & - & - & 42.3 & - & -  
    & 48.5 & 69.8 & 53.2 & 43.4 & 66.8 & 46.9 \\
    MPViT-B \cite{lee2021mpvit} & 95 & 503
    & -    & -    & -    & -    & -     & -  
    & 49.5 & 70.9 & 54.0 & \textbf{44.5} & 68.3 & \textbf{48.3} \\
    Focal-B \cite{yang2021focal} & 110 & 533 
    & \textbf{47.8} & - & - & 43.2 & - & - 
    & 49.0 & 70.1 & 53.6 & 43.7 & 67.6 & 47.0  \\
    \rowcolor{gray!20} Dilate-B  (ours) & 67 & 370
    &  47.6 & \textbf{70.2} &	\textbf{55.2} & \textbf{43.4} & \textbf{67.2} & \textbf{46.8}
    & \textbf{49.9} & \textbf{71.9} & \textbf{55.1} & \textbf{44.5} & \textbf{68.9} & 47.7 \\
    \bottomrule
    \end{tabular}
    \vspace{-1em}
    \label{coco_mrcnn_tab}
\end{table*}

\subsection{Object Detection and Instance Segmentation on COCO}
\label{detect}
\noindent \textbf{- Dataset and implementation details.} 
We evaluate our variants on object detection and instance segmentation on COCO2017 dataset \cite{lin2014microsoft}. COCO2017 dataset contains 118K images for training, 5K images for validation and 20K images for testing.
We utilize two representative frameworks: Mask R-CNN \cite{he2017mask} and Cascade Mask R-CNN \cite{cai2019cascade} implemented in mmdetection  \cite{chen2019mmdetection} and adopt the ImageNet-1K pre-trained variants as backbones. For Mask R-CNN and Cascade Mask R-CNN frameworks, we use the AdamW optimizer with a base learning rate of 0.0001, a weight decay of 0.05, and a batch size of 16. For a fair comparison, we train our variants Dilate-S and Dilate-B via two strategies: 
(1) 1× schedule with 12 epochs where the shorter side
of the image is resized to 800 and the longer side is less than 1333;  
(2) 3× schedule with 36 epochs where the multi-scale training strategy is adopted and the shorter side of the image is resized in $[480, 800]$. Because image resolution in object detection and instance segmentation is generally larger than that in image classification, we use a combination of local window attention, local window attention with shifted operation \cite{liu2021swin} and global attention in stage3 of DilateFormer to reduce computational cost.

\vspace{0.1cm}

\noindent  \textbf{- Results and analysis.}
Table \ref{coco_mrcnn_tab} and Table \ref{coco_cascsde_tab} report box mAP ($\rm AP^b$) and mask mAP ($\rm AP^m$) of Mask R-CNN framework and Cascade Mask R-CNN framework, respectively. Our DilateFormer variants outperform recent Transformers on both object detection and instance segmentation in two frameworks. For Mask R-CNN $1\times$ schedule, DilateFormer surpasses Swin Transformer \cite{liu2021swin} by 2.8-3.6\% of box mAP and 2.5-2.6\% mask mAP at comparable settings, respectively. For $3\times$ + MS schedule, Dilate-B achieves 49.9\% box mAP and 43.7\% mask mAP in Mask R-CNN framework, 53.3\% box mAP and 46.1\% mask mAP in Cascade Mask R-CNN framework.
Furthermore, our Dilate-S outperforms PVT-M \cite{wang2021pyramid}  by 2.2\% box mAP, 2.7\% mask mAP at $1 \times$ schedule with 13.2\% fewer FLOPs.

\subsection{Semantic Segmentation on ADE20K}
\label{seg}
\noindent  \textbf{- Dataset and implementation details.}
ADE20K dataset \cite{zhou2017scene} contains  150 semantic categories, and there are 20,000 images for training, 2000 images for validation and 3000 images for testing.
We evaluate the proposed variants for DilateFormer on semantic segmentation on ADE20K and utilize two representative frameworks: Upernet \cite{xiao2018unified} and Semantic FPN \cite{kirillov2019panoptic} implemented in mmsegmentation \cite{mmseg2020} with our ImageNet-1K pre-trained variants as backbones. For training Upernet, we follow the configuration of Swin Transformer and train our variants for 160K iterations.  We employ the AdamW \cite{Loshchilov2019adamw} optimizer with a base learning rate of 0.00006, a weight decay of 0.01, a batch size of 16, and a linear scheduler with a linear warmup of 1,500 iterations. As for Semantic FPN 80K iterations, we follow the same configuration of PVT with a cosine learning rate schedule with an initial learning rate of 0.0002 and a weight decay of 0.0001. 

\begin{table*}[h]
    \centering
    \small
    \captionsetup{justification=centering}
    \caption{ \textsc{Object detection and instance segmentation with Cascade Mask R-CNN on COCO val2017.}}
    \vspace{-0.5em}
    \begin{tabular}{c|c|c|c|c|c|c|c|c}
    \toprule
    \multirow{2}{*}{Method} 
    & Params & FLOPs & \multicolumn{6}{c}{$3\times$ + MS schedule} \\
    & (M) & (G) & {AP}$^b$ & {AP}$^{b}_{50}$ & {AP}$^{b}_{75}$ & {AP}$^{m}$  & {AP}$^{m}_{50}$  & {AP}$^{m}_{75}$ \\
    \midrule
    Res50 \cite{resnet}    & 82 & 739 & 46.3 & 64.3 & 50.5 & 40.1 & 61.7 & 43.4 \\
    NAT-T \cite{Hassani2022nat} & 85 & 737 & 51.4 & 70.0 & 55.9 & 44.5 & 67.6 & 47.9 \\
    ConvNeXt-T \cite{liu2022ConvNeXt}  & 86 & 741 & 50.4 & 69.1 & 54.8 & 43.7 & 66.5 & 47.3 \\
    Swin-T \cite{liu2021swin}   & 86 & 745 & 50.5 & 69.3 & 54.9 & 43.7 & 66.6 & 47.1 \\
    Shuffle-T  \cite{Huang2021shuffle} & 86 & 746 & 50.8 & 69.6 & 55.1 & 44.1 & 66.9 & 48.0 \\
    UniFormer-S$_{h14}$ & 79 & 747 & 52.1 & 71.1 & 56.6 & 45.2 & 68.3 & 48.9 \\
    DeiT-S \cite{deit}    & 80 & 889 & 48.0 & 67.2 & 51.7 & 41.4 & 64.2 & 44.3 \\
    \rowcolor{gray!20} Dilate-S  (ours) & 82 & 740 & \textbf{52.4} & \textbf{71.6} &	\textbf{56.9} & \textbf{45.2} & \textbf{68.6} & \textbf{49.0} \\ 
    \midrule
    X101-32 \cite{xie2017AgRTr} & 101 & 819 & 48.1 & 66.5 & 52.4 & 41.6 & 63.9 & 45.2 \\
    NAT-S \cite{Hassani2022nat}   & 108 & 809 & 52.0 & 70.4 & 56.3 & 44.9 & 68.1 & 48.6 \\
    ConvNeXt-S \cite{liu2022ConvNeXt} & 108 & 827 & 51.9 & 70.8 & 56.5 & 45.0 & 68.4 & 49.1 \\
    Swin-S \cite{liu2021swin}   & 107 & 838 & 51.8 & 70.4 & 56.3 & 44.7 & 67.9 & 48.5 \\
    NAT-B \cite{Hassani2022nat}     & 147 & 931 & 52.3 & 70.9 & 56.9 & 45.1 & 68.3 & 49.1 \\
    ConvNeXt-B \cite{liu2022ConvNeXt} & 146 & 964 & 52.7 & 71.3 & 57.2 & 45.6 & 68.9 & 49.5 \\
    Swin-B \cite{liu2021swin}  & 145 & 982 & 51.9 & 70.5 & 56.4 & 45.0 & 68.1 & 48.9 \\
    \rowcolor{gray!20} Dilate-B  (ours) & 105 & 849 & \textbf{53.5} & \textbf{72.4} & \textbf{58.0} & \textbf{46.1} & \textbf{69.9} & \textbf{50.3} \\
    \bottomrule
    \end{tabular}
    \vspace{1em}
   \label{coco_cascsde_tab}
\end{table*}

\vspace{0.1cm}

\noindent  \textbf{- Results and analysis.} Table \ref{tab:seg_tab} shows the results of DilateFormer equipped with UperNet and Semantic FPN frameworks. Our variants DilateFormer-Small/Base equipped with UperNet framework achieve 47.1/50.4\% mIoU and 47.6/50.5\% MS mIoU, outperforming Swin \cite{liu2021swin} by at least 2.6\% of mIoU and 1.0\% of MS mIoU respectively. For Semantic FPN framework, our variants achieve 47.1/48.8\% mIoU, and exceed Swin \cite{liu2021swin}
by 3.6-5.6\%.

\begin{table*}[t]
\setlength{\tabcolsep}{6pt} 
\renewcommand{\arraystretch}{1} 
    \captionsetup{justification=centering}
    \caption{\textsc{Semantic segmentation experimental results on ADE20K validation set. \\
    Left:  with Upernet; Right:  with semantic FPN.}}
    \vspace{-0.5em}
    \centering
      \small
\begin{tabular}{c|cccc}
            \toprule
            \multirow{3}{*}{Method} 
            & \multicolumn{4}{c}{Upernet 160K} \\
            & Params & FLOPs & mIoU & MS mIoU \\
            & (M)    & (G)   & (\%) & (\%)\\
            \midrule
            Res101 \cite{resnet}  & 86 & 1029 & -    & 44.9 \\
            Twins-S \cite{chu2021twins} & 54 & 901 & 46.2 & 47.1 \\
            TwinsP-S \cite{chu2021twins} & 55 & 919 & 46.2 & 47.5 \\
            ConvNeXt-T \cite{liu2022ConvNeXt} & 60 & 939 & 46.0 & 46.7 \\
            TRT-ViT-B \cite{xia2022trt} & 81 & 941 & 46.5 & 47.5 \\
            GG-T \cite{yu2021glance} & 60 & 942 & 46.4 & 47.2 \\
            Swin-T \cite{liu2021swin}  & 60 & 945 & 44.5 & 45.8 \\
            Shuffle-T \cite{Huang2021shuffle} & 60 & 949 & 46.6 & \textbf{47.8} \\
            Focal-T \cite{yang2021focal} & 62 & 998 & 45.8 & 47.0 \\
            \rowcolor{gray!20} Dilate-S  (ours) & 54 & 935 & \textbf{47.1} & 47.6 \\
            \midrule
            NAT-S \cite{Hassani2022nat}     & 82 & 1010 &  48.0 & 49.5 \\
            Twins-B \cite{chu2021twins} & 89 & 1020 & 47.7 & 48.9 \\
            ConvNeXt-S \cite{liu2022ConvNeXt} & 82 & 1027 & 48.7 & 49.6 \\
            Swin-S \cite{liu2021swin}  & 81 & 1038 & 47.6 & 49.5 \\
            GG-S  \cite{yu2021glance} & 81 & 1035 & 48.4 & 49.6 \\
            TRT-ViT-D \cite{xia2022trt} & 144 & 1065 & 48.8 & 49.8 \\
            Shuffle-B  \cite{Huang2021shuffle} & 121 & 1196 & 49.0 & 50.5 \\
            Next-ViT-L \cite{li2022next} & 92 & 1072 & 50.1 & 50.8 \\
            Focal-S \cite{yang2021focal}  & 85 & 1130 & 48.0 & 50.0 \\
            \rowcolor{gray!20} Dilate-B  (ours) & 79 & 1046 & \textbf{50.8} & \textbf{51.1} \\
            \bottomrule
\end{tabular}
        \label{seg_upernet_tab}
    \hspace{30pt}
      \centering
      \small
\begin{tabular}{c|ccc}
            \toprule
            \multirow{3}{*}{Method} 
            & \multicolumn{3}{c}{Semantic FPN 80K} \\
            & Params & FLOPs & mIoU \\
            & (M)    & (G)   & (\%) \\
            \midrule
            Res50 \cite{resnet}   & 29 & 183 & 36.7 \\
            Twins-S \cite{chu2021twins}  & 28 & 144 & 43.2 \\
            PVT-S \cite{wang2021pyramid}    & 28 & 161 & 39.8 \\
            TwinsP-S \cite{chu2021twins} & 28 & 162 & 44.3 \\
            XCiT-S12/8 \cite{ali2021XCiT} & 30 & - &  44.2 \\
            TRT-ViT-B \cite{xia2022trt} & 46 & 176 & 45.4 \\
            Swin-T \cite{liu2021swin}  & 32 & 182 & 41.5 \\
            Next-ViT-S \cite{li2022next} & 36 & 208 & 46.5 \\ 
            CrossFormer-S \cite{Wang2021crossf} &34 & 209 & 46.4 \\
            \rowcolor{gray!20} Dilate-S  (ours) & 28 & 178 & \textbf{47.1} \\
            \midrule
            Res101 \cite{resnet}  & 48 & 260 & 38.8 \\
            Next-ViT-B \cite{li2022next} & 49 & 260 & 48.6 \\
            XCiT-S24/8 \cite{ali2021XCiT} & 52 & - &  47.1 \\
            Swin-S \cite{liu2021swin}  & 53 & 274 & 45.2 \\
            PVT-L \cite{wang2021pyramid}    & 65 & 283 & 42.1 \\
            TwinsP-L \cite{chu2021twins} & 65 & 283 & 46.4 \\
            TRT-ViT-D \cite{xia2022trt} & 106 &  296 & 46.7 \\
            CrossFormer-B \cite{Wang2021crossf} & 56 & 320 & 48.0 \\
            Swin-B \cite{liu2021swin}  & 91 & 422 & 46.0 \\
            \rowcolor{gray!20} Dilate-B  (ours) & 51 & 288 & \textbf{48.8} \\
            \bottomrule
\end{tabular}
        \label{seg_fpn_tab}     
    \vspace{-1.5em}
    \label{tab:seg_tab}
\end{table*}

\subsection{Ablation Studies}
\label{ablate_study}
We conduct ablation studies from the perspectives of sparse and local patterns, dilation scale, block setting, stage setting and overlapping tokenizer/downsampler. More ablation studies about the kernel size are given in the supplementary material.

\vspace{0.1cm}

\noindent  \textbf{- SWDA vs. other sparse and local patterns.}
We replace Sliding Window Dilated Attention (SWDA) in the first two stages with other sparse and local patterns, \ie, Dilated Convolution (DC) \cite{yu2015multi}, Dynamic Dilated Convolution (DDC) \cite{chen2020dynamic} , Window Attention with Spatial Shuffle (WASS\protect\footnotemark[2]) \cite{Huang2021shuffle} and Sliding Window Attention (SWA) \cite{li2021simvit}. 

\stepcounter{footnote}\footnotetext{The WASS is an approximate sparse sampling operation which divides patches into local Windows like Swin \cite{liu2021swin} and then shuffles keys and values between different windows.}

As shown in Table \ref{tab:other_local_sparse}, our SWDA outperforms other sparse and local patterns in various vision tasks. SWDA achieves 82.1\% Top-1 accuracy on ImageNet-1K, 44.9\% box mAP/40.9\% mask mAP on COCO and 45.84\% mIoU on ADE20K. SWDA outperforms DC (+0.4\%, +1.4\%/+0.6\%, +1.69\%) because attention is data-specific compared to conventional convolution. Although DDC is local, sparse and data-specific like SWDA, SWDA still outperforms DDC (+0.3\%, +0.6\%/+0.3\%, +0.94\%). DDC uses the entire feature map to generate the kernel parameter of convolution, which is data-specific at the feature-map level; and in comparison, SWDA performs self-attention on keys and values sparsely selected in a sliding window centered on the query patch, which is data-specific at the token level. Therefore, SWDA has a stronger modeling capability than DDC. SWDA also outperforms WASS (+0.3\%, +0.8\%/+0.5\%, +1.18\%) and SWA (+0.3\%, +0.5\%/+0.1\%, +2.21\%), which demonstrates the importance of considering locality and sparsity in self-attention of the shallow layers.

\begin{table}[t]
    \centering
    \renewcommand\arraystretch{1.1}
    \captionsetup{justification=centering}
    \caption{\textsc{Experiment results with local and sparse patterns in the first two stages. The} Top-1 \textsc{is on ImageNet-1K, AP$^{b}$ and AP$^{m}$ are on COCO val2017 with Mask R-CNN 1× schedule,} mIoU \textsc{is on ADE20K validation set with semantic FPN.}}
    \vspace{-0.5em}
    \resizebox{\linewidth}{!}{
    \begin{tabular}{c|c|c|c|c|c|c|c}
    \toprule
    \multirow{2}{*}{Pattern} & \multirow{2}{*}{Locality} & \multirow{2}{*}{Sparity} & Data  & Top-1 &  {AP}$^b$ &  {AP}$^m$   & mIoU \\
    & & & Specific  & (\%) & (\%) & (\%) & (\%)  \\
    \midrule
    DC & $\checkmark$ & $\checkmark$ & &  81.7 & 43.5 & 40.3 & 44.15 \\
    DDC & $\checkmark$ & $\checkmark$ & $\checkmark$ & 81.8	& 44.3 & 40.6 & 44.90 \\
    WASS &  & $\checkmark$ & $\checkmark$ & 81.8 & 44.1 & 40.4 & 44.66 \\
    SWA & $\checkmark$   &  & $\checkmark$ & 81.8	& 44.4 & 40.8 & 43.63 \\
    \rowcolor{gray!20} SWDA & $\checkmark$ & $\checkmark$ & $\checkmark$& \textbf{82.1}	& \textbf{44.9} & \textbf{40.9} & \textbf{45.84} \\
    \bottomrule
    \end{tabular}
    }
    \vspace{0.5em}
    \label{tab:other_local_sparse}
\end{table}

\begin{table}[t]
\renewcommand\arraystretch{1.1}
\makeatletter\def\@captype{table}
\centering
\captionsetup{justification=centering}
\caption{\textsc{Top-1 accuracy on ImageNet-1K of different dilation scales.}}
\begin{tabular}{c|c|c|c}
    \toprule
    \multirow[c]{2}{*}{Scale} & Head Num       &Dilation  & Top-1  \\
     & in Stage1/2    &Rate      &  (\%)\\
    \midrule
    \multirow{5}{*}{Multi-} &$[2,~4]$   & $[1,~2]$         &  81.7 \\
    \cline{2-4}
    & \multirow{3}{*}{$[3,~6]$}   & $[1,~2,~3]$      &  \textbf{82.1} \\
    & ~         & $[2,~3,~4]$      &  81.8 \\ 
    & ~         & $[3,~4,~5]$      &  81.7 \\
    \cline{2-4}
    &$[4,~8]$  & $[1,~2,~3,~4]$   &  81.9 \\
    \cline{1-4}
    \multirow{3}{*}{Single}& \multirow{3}{*}{$[3,~6]$}  & $[1]$      &  81.7     \\ 
    &   & $[2]$      &  81.9     \\
    &   & $[3]$      &  81.8     \\ 
    \bottomrule
\end{tabular}
\vspace{-1.5em}
\label{dilate_scale_tab}
\end{table}

\noindent \textbf{- Dilation scale.}
\label{dilation_scale}
Since the number of heads must be multiple of the number of dilation scales, we change the number of heads and feature dimensions in each head, keeping the same total length according to the number of dilation scales. We analyze the effects of dilation scales according to the performance on ImageNet-1K classification task. The number of heads in stage 1 or 2, dilation scales and top-1 accuracy are shown in Table \ref{dilate_scale_tab}. With the same number of heads in the block, the top-1 accuracy (82.1\%) of multi-scale dilated attention, \ie, $[1,~2,~3]$, is better than that of single-scale, \ie, $[1]$, $[2]$, and $[3]$, because multi-scale can provide richer information than single-scale. What is more, the dilation rates in the block need to be moderate so that it can simultaneously model both locality and sparsity of attention, without introducing redundant information modeling due to the large receptive field such as global attention. Therefore, we set the dilation scale of the model to 3, \ie, $[1,~2,~3]$ by default.

\vspace{0.1cm}

\noindent  \textbf{- MSDA vs. other blocks setting.} In our DilateFormer, we stack Multi-Scale Dilated Attention (MSDA) blocks in the first two stages. To demonstrate the effectiveness of our proposed MSDA, we replace MSDA in the first two stages of the default setting (D-D-G-G) with local attention in a shifted window (L-L-G-G) \cite{liu2021swin} and global attention (G-G-G-G) \cite{dosovitskiy2020image} for comparison. We also compare with the global attention cooperated with a naïve downsampling technique, namely global attention with spatial reduction (G-G-G-G + sr.) \cite{wang2021pyramid}, which reduces the redundant interaction between patches by decreasing the number of patches. The maximum size of attended receptive field in MSDA is $7\times7$ with dilation, the size of attended receptive field in local attention is $7\times7$, and the size of attended receptive field in global attention is the size of the entire feature map.

\vspace{0.1cm}
Table \ref{blocks_setting_table} summarizes the comparison results. By using the same size of maximum attended receptive field, our MSDA (82.1\%) outperforms local attention with shifted window (L-L-G-G) \cite{liu2021swin} (81.7\%) with fewer FLOPs, which demonstrates the effectiveness of sparse and local attention mechanisms in shallow layers. Compared with the global attention (G-G-G-G) \cite{dosovitskiy2020image}, our MSDA achieves an improvement of 0.3\% with half of FLOPs, which further demonstrates the effectiveness and efficiency of the proposed local and sparse attention mechanism. Also, the superiority of MSDA against the global attention shows the redundancy of modeling dependencies among all image patches. To reduce the redundant interaction, the global attention with spatial reduction utilizes downsampling by convolution but introduces extra parameters. By contrast, our MSDA exploits the locality and sparsity property without extra parameters. The results show that our MSDA surpasses the global attention with spatial reduction by 0.5\%, which indicates the effectiveness of redundancy reduction of the proposed MSDA. In downstream tasks, our MSDA block also outperforms other types of blocks, indicating that MSDA has a stronger modeling capability.

\begin{table}[t]
    \centering
    \renewcommand\arraystretch{1.2}
    \captionsetup{justification=centering}
    \caption{\textsc{Experiment results with different blocks  in Stage1/2.} ``sr.'' \textsc{indicates spatial reduction  operation. ``D'', ``G'' and ``L'' indicate dilation, global and local operations, respectively. The} Top-1 \textsc{ is on ImageNet-1K, AP$^{b}$ and AP$^{m}$ are on COCO val2017 with Mask R-CNN 1 × schedule,} mIoU \textsc{is on ADE20K validation set with semantic FPN.}}
    \vspace{-0.5em}
   \begin{tabular}{c|c|c|c|c|c|c}
    \toprule
     \makecell[c]{Block \\ Type}  & \makecell[c]{Params \\ (M) } & \makecell[c]{FLOPs \\ (G) } & \makecell[c]{Top-1 \\ (\%)} & \makecell[c]{{AP}$^b$ \\ (\%)} & \makecell[c]{{AP}$^{m}$ \\ (\%)} & \makecell[c]{mIoU \\ (\%)}  \\   
    \midrule
    G-G-G-G~+~sr. & 20.6 & 3.02 & 81.6 & 40.9 & 37.9	& 44.4 \\
    G-G-G-G & 17.2 & 6.36 & 81.8 & 42.0 & 38.7 & 44.5 \\
     L-L-G-G & 17.2 & 3.24 & 81.7 & 40.9 & 37.6 & 44.3 \\
    \rowcolor{gray!20} D-D-G-G      & 17.2 & 3.18 & \textbf{82.1} & \textbf{44.2} & \textbf{40.9} & \textbf{45.8} \\
    \bottomrule
\end{tabular}
    \label{blocks_setting_table}
    \vspace{0.5em}
\end{table}

\begin{table}[t]
    \centering
    \captionsetup{justification=centering}
    \caption{\textsc{ Analysis of Multi-Scale Dilated Attention blocks in different stages on ImageNet-1K.}}
    \begin{tabular}{c|c|c}
    \toprule
     Stage    & FLOPs & Top-1 \\ 
     Setting  & (G)   & (\%)  \\
    \midrule
    G-G-G-G  & 6.36 & 81.8 \\
    D-G-G-G  & 3.53 & 82.2 \\
    \rowcolor{gray!20} D-D-G-G  & 3.18 & 
    \textbf{82.1} \\
    D-D-D-G  & 3.05 & 81.3 \\
    D-D-D-D  & 3.04 & 80.5 \\
    \bottomrule
    \end{tabular}
    \vspace{-1.5em}
    \label{stages_setting_table}
\end{table}

\begin{table}[t]
    \centering
    \renewcommand\arraystretch{1.2}
    \captionsetup{justification=centering}
    \caption{Top-1 \textsc{accuracy on ImageNet-1K of using Overlapping Tokenizer and Downsampler.}}
    \vspace{-0.5em}
    \begin{tabular}{c|c|c|c|c}
    \toprule
    Overlapping  & Overlapping & Params & FLOPs & Top-1 \\ 
    Tokenizer    & Downsampler & (M)  & (G)   & (\%)  \\
    \midrule
    ~ & ~ & 16.1 & 2.62 & 81.7 \\
    ~ & $\checkmark$ & 17.2 & 2.74 & 81.8 \\
    $\checkmark$ & ~ & 16.2 & 3.12 & 81.9 \\
    \rowcolor{gray!20} $\checkmark$ & $\checkmark$ & 17.2 & 3.18 & \textbf{82.1} \\
    \bottomrule
    \end{tabular}
    \vspace{0.5em}
    \label{overlap_table}
\end{table}

\vspace{0.1cm}
\noindent  \textbf{- Stage setting.}
To demonstrate the modeling capability of the MSDA block at shallow stages, we conduct a set of experiments to explore the performance of using MSDA in different stages.
In the four stages of the model, we progressively replace the global MHSA block in each stage with the MSDA block. Table \ref{stages_setting_table} shows FLOPs and top-1 accuracy of models with different structures. 
The model performance shows a decreasing trend, from 82.2\% down to 80.5\%, as the proportion of MSDA blocks in the model stage increases. 
The results show that it is more effective to consider the locality and sparsity property of the self-attention mechanism in shallow stages rather than in deeper stages. What's more, the model with MSDA block only in stage1 (82.2\%) performs slightly better than the model with MSDA blocks in both stage1 and stage2 (82.1\%), but the former has larger FLOPs (+ 0.35G). Therefore, we use MSDA blocks in both stage1 and stage2 by default.

\begin{table}[t]
\renewcommand\arraystretch{1.2}
\makeatletter\def\@captype{table}
\centering
\captionsetup{justification=centering}
\caption{\textsc{Comparison of model inference. “Mem” denotes the peak memory for evaluation. “FPS” is the number of images processed for one second.}}
\vspace{-0.5em}
\resizebox{\linewidth}{!}{
\begin{tabular}{c|c|c|c|c|c}
\toprule
\multirow{2}{*}{Method} & Params & FLOPs & FPS & Mem. & Top-1 \\
&      (M)  &  (G)  &   (s)   &   (G)    &  (\%)   \\
\midrule
ConvNeXt-T \cite{liu2022ConvNeXt} & 28     & 4.5   & 2450  & 3.5  & 82.1  \\
Swin-T \cite{liu2021swin}    & 28     & 4.5   & 1681  & 5.0  & 81.3  \\
NAT-T \cite{Hassani2022nat}      & 28     & 4.5   & 1515  & 3.7  & 83.2  \\
DiNAT-T \cite{hassani2022dilated}   & 28     & 4.5   & 1479  & 3.7  & 82.7  \\
\rowcolor{gray!20} Dilate-S (ours)   & 21     & 4.8   & 1368  & 3.1  & \textbf{83.3}  \\
\midrule
ConvNeXt-S \cite{liu2022ConvNeXt} & 50     & 8.7   & 1558  & 4.8  & 83.1  \\
Swin-S \cite{liu2021swin}     & 50     & 8.7   & 1045  & 6.7  & 83.0  \\
NAT-S \cite{Hassani2022nat}     & 51     & 7.8   & 1055  & 5.0  & 83.7  \\
DiNAT-S \cite{hassani2022dilated}   & 51     & 7.8   & 1069  & 5.0  & 83.8  \\
\rowcolor{gray!20} Dilate-B (ours)   & 47     & 10.0  & 1122  & 4.0  & \textbf{84.4}  \\
\bottomrule
\end{tabular}
}
\vspace{-2em}
\label{tab:fps}
\end{table}

\vspace{0.1cm} 
\noindent  \textbf{- Overlapping Tokenizer/Downsampler.}
\label{overlapping}
We further study how the overlapping tokenizer or downsampler affect the performance.
While keeping the same settings, we replace our overlapping tokenizer or downsampler with a simple non-overlapping tokenizer or downsampler, i.e., convolution with kernel size 4 and stride 4 or convolution with kernel size 2 and stride 2. As shown in Table \ref{overlap_table}, our model achieves a slight improvement (+0.4\%) with overlapping tokenizer/downsampler, indicating that the main improvement of our model does not rely on these two modules.
\vspace{0.1cm}

\noindent \textbf{- Comparisons of real running times.} We provide a comparison of model inference about FPS, peak memory about our DilateFormers and current SOTA models in Table \ref{tab:fps}. FPS and peak memory usage are measured from forward passes with a batch size of 256 on a single A100 GPU. With comparable model parameters and FLOPs, our DilateFormers have comparable FPS and better performance than current SOTA models.

\vspace{0.1cm} 
\noindent   \textbf{- Grad-CAM Visualization.}
\label{grad_cam}
To further illustrate the recognition ability of the proposed DilateFormer, we apply Grad-CAM \cite{jacobgilpytorchcam}  to visualize the areas of the greatest concern in the last layer of DeiT-Tiny \cite{deit}, Swin-Tiny \cite{liu2021swin} and Dilate-Tiny.  As shown in Figure \ref{fig:grad_cam}, our Dilate-Tiny model performs better in locating the target objects and attends to semantic areas more continuously and completely, suggesting the stronger recognition ability of our model. Such ability yields better classification performance compared with DeiT-Tiny and Swin-Tiny.

\vspace{0.1cm} 
\noindent  \textbf{- More Visualization Results on Global Attention.}
\label{more_attn_results}
In Sec.\ref{intro}, we discuss two key properties \ie, \textit{locality} and \textit{sparsity} of global attention in shallow layers. To further analyze these two properties, we visualize more attention maps in the shallow layers of ViT-Small \cite{dosovitskiy2020image}. As shown in Figure \ref{fig:more_atten}, the attention maps in the shallow layers of ViT-Small show that activated key patches are sparsely distributed in the neighborhood of the query patch. Specifically, the patches with high attention scores sparsely scatter around the query patch and other patches have low attention scores.

\begin{figure}[t]
    \centering
    \vspace{-2em}
    \includegraphics[width=\linewidth]{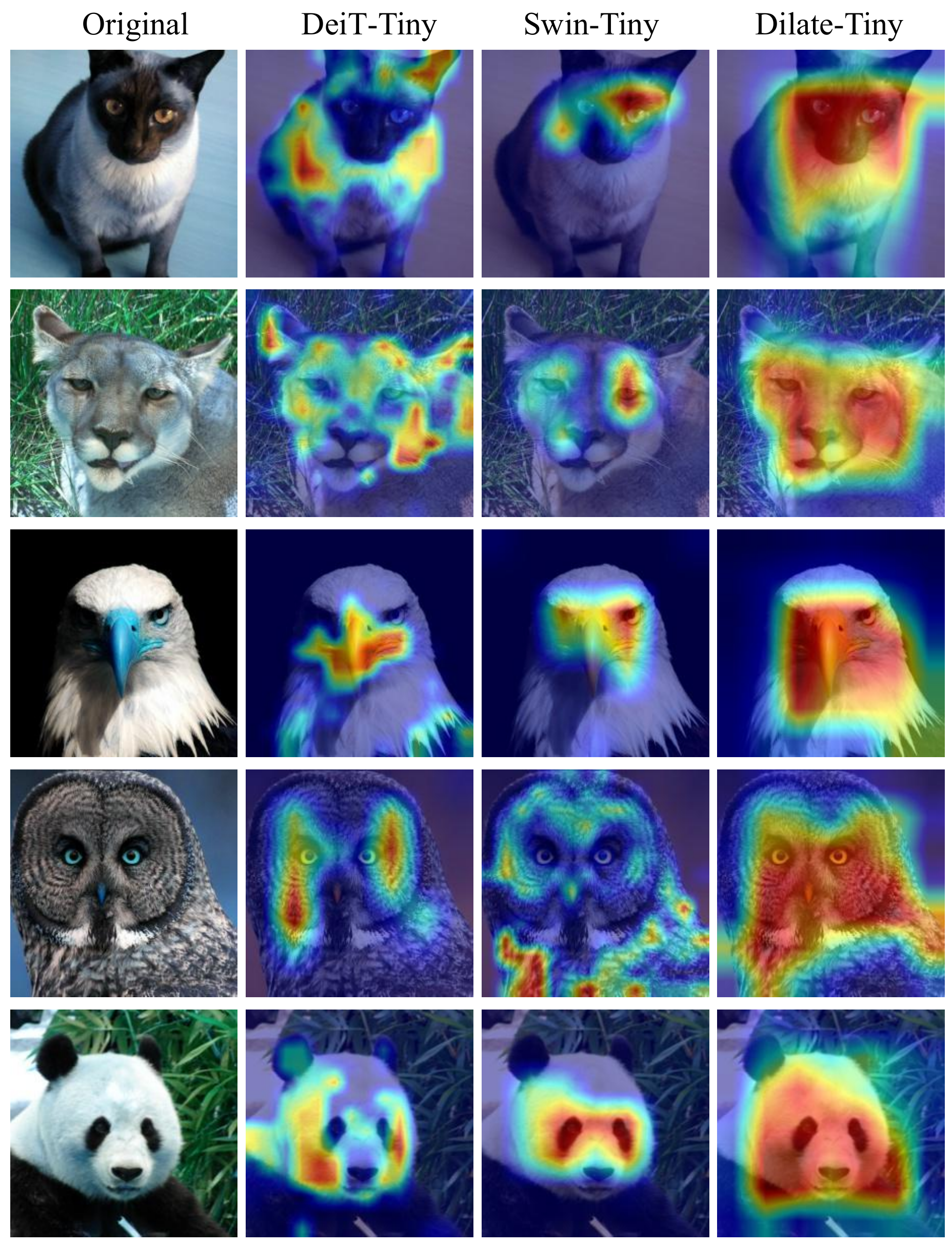}
    \caption{Grad-CAM Visualization of the last layer of DeiT-Tiny, Swin-Tiny and Dilate-Tiny. Images are from the validation set of ImageNet-1k.}
    \vspace{-1.5em}
    \label{fig:grad_cam}
\end{figure}

\section{Conclusion}
\label{conclude}
In this work, we propose a strong and effective Vision Transformer, called DilateFormer, which can provide powerful and general representations for various vision tasks.
Our proposed Multi-Scale Dilated Attention (MSDA) takes both the locality and sparsity of the self-attention mechanism in the shallow layers into consideration, which can effectively aggregate semantic multi-scale information and efficiently reduce the redundancy of the self-attention mechanism without complex operations and extra computational cost.
Extensive experiment results show that the proposed method achieves state-of-the-art performance in both ImageNet-1k classification and down-stream vision tasks such as object detection and semantic segmentation.

\begin{figure}[t]
    \centering
    \includegraphics[width=\linewidth]{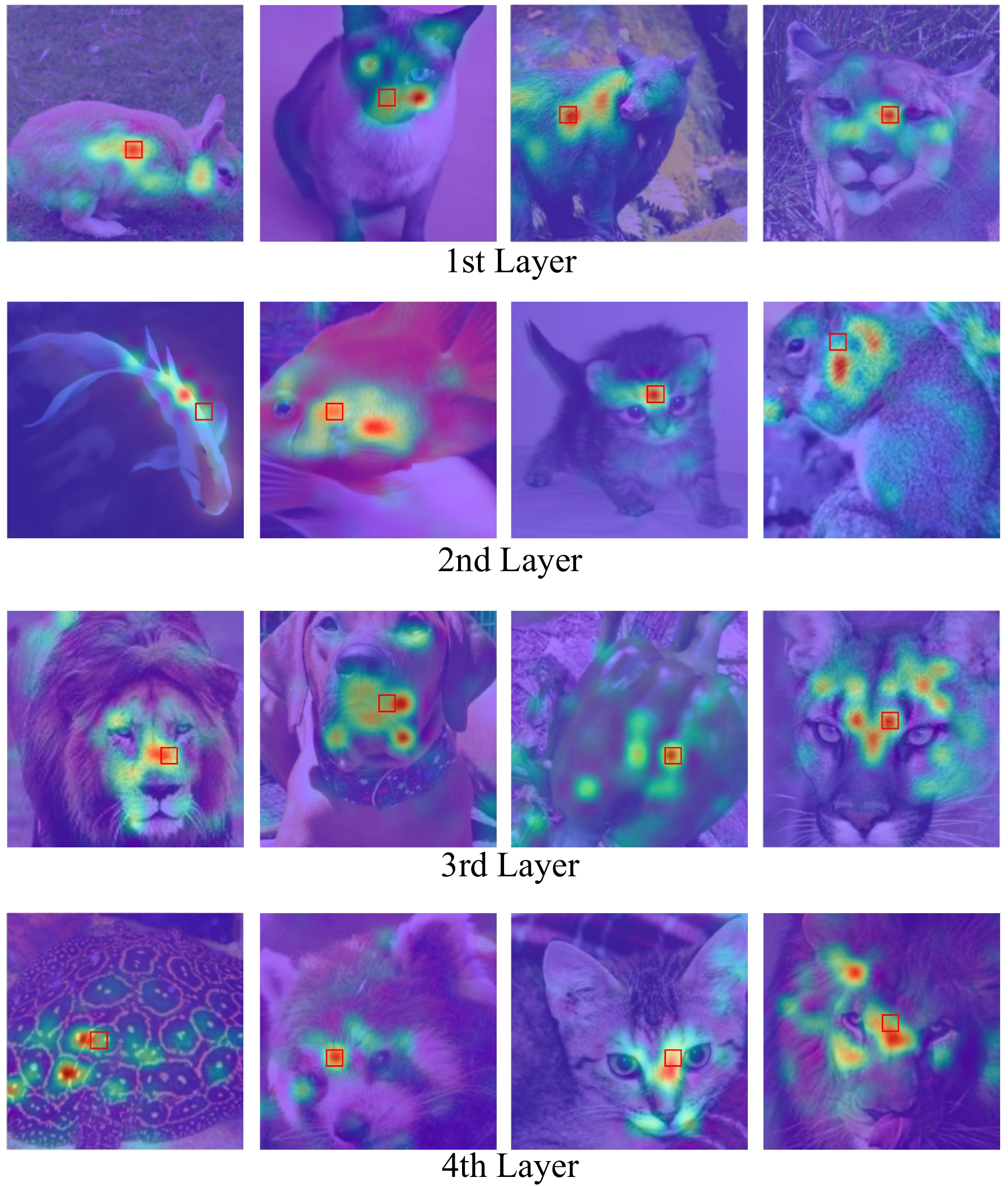}
    \caption{More Visualization of attention maps of shallow layers of ViT-Small. We visualize the activations in attention maps of the query patches (in the red box). The attention maps show that patches with high attention scores sparsely scatter around the query patch, and other patches have low attention scores.}
    \vspace{-1.5em}
    \label{fig:more_atten}
\end{figure}

\section*{Acknowledgments}
This work was supported partially by the NSFC (U21A20471,U1911401,U1811461), Guangdong NSF Project (No. 2020B1515120085, 2018B030312002), Guangzhou Research Project (201902010037), and the Key-Area Research and Development Program of Guangzhou (202007030004).

\bibliographystyle{ieee_fullname}
\normalem
\bibliography{citation}

\newpage
\textbf{\huge{Supplementary Material}}
\section*{More ablation studies}
\label{abs}

\noindent \textbf{-Minimum Kernel size.}
\label{window_size}
The minimum kernel size affects the size of the attended receptive field of our Sliding Window Dilated Attention operation and naturally affects the performance of our DilateFormer. We analyze the effect of different minimum kernel sizes on the ImageNet-1K classification task. The experimental results in terms of FLOPs and Top-1 accuracy are shown in Table \ref{kernel_tab}. Experiments show that using a larger kernel size in SWDA can slightly improve the performance due to a larger attended receptive field, but the FLOPs will also increase. By default, we set the kernel size in SWDA to $3\times3$. 

\begin{table}[h]
\small
\centering
\makeatletter\def\@captype{table}
\captionsetup{justification=centering}
\caption{\textsc{ Top-1 accuracy on ImageNet-1K of  different  minimum   kernel sizes.}}
\vspace{-0.5em}
\begin{tabular}{c|c|c}
    \toprule
    Minimum  & FLOPs & Top-1  \\
    Kernel Size   &  (G)  &  (\%)  \\
    \midrule
    $3\times3$   &  3.18  & 82.1 \\
    $5\times5$   &  3.21  & 82.2 \\ 
    $7\times7$   &  3.24  & 82.3 \\
    \bottomrule
\end{tabular}
\label{kernel_tab}
\end{table}

}
\end{document}